\documentclass{article}
\usepackage[T1]{fontenc}
\usepackage[utf8]{inputenc}
\usepackage{graphicx}
\usepackage{booktabs}
\usepackage[english]{babel}
\usepackage{xspace}
\usepackage{todonotes}
\usepackage{longtable}
\usepackage{array}
\usepackage{booktabs}
\usepackage{subcaption}
\usepackage{caption}
\usepackage{lscape}
\usepackage{geometry}
\usepackage{amsmath}
\usepackage[normalem]{ulem}

\usepackage{url}
\usepackage[colorlinks=true, allcolors=blue]{hyperref}

\title{Data Processing for the OpenGPT-X Model Family}
\author{
Nicolo' Brandizzi$^1$, 
Hammam Abdelwahab$^1$, 
Anirban Bhowmick$^1$, \\
Lennard Helmer$^1$, 
Benny Jörg Stein$^1$, 
Pavel Denisov$^1$, 
Qasid Saleem$^1$, \\
Michael Fromm$^1$, 
Mehdi Ali$^1$,
Richard Rutmann$^1$, \\
Farzad Naderi$^2$, 
Mohamad Saif Agy$^2$, 
Alexander Schwirjow$^2$, 
Fabian Küch$^2$,\\ 
Luzian Hahn$^2$, 
Malte Ostendorff$^3$, 
Pedro Ortiz Suarez$^3$, 
Georg Rehm$^3$,\\ 
Dennis Wegener$^1$,
Nicolas Flores-Herr$^1$,
Joachim Köhler$^1$,
Johannes Leveling$^1$\\\\
$^1$ Fraunhofer IAIS,  <firstname>.<lastname>@iais.fraunhofer.de\\
$^2$ Fraunhofer IIS,  <firstname>.<lastname>@iis.fraunhofer.de\\
$^3$ DFKI,  <firstname>.<lastname>@dfki.de
}
\date{}
\begin{document}
\maketitle

\begin{abstract}
This paper presents a comprehensive overview of the data preparation pipeline developed for the OpenGPT-X project, a large-scale initiative aimed at creating open and high-performance multilingual large language models (LLMs). The project goal is to deliver models that cover all major European languages, with a particular focus on real-world applications within the European Union. 
We explain all data processing steps, starting with the data selection and requirement definition. We distinguish between curated data and web data, as each of these categories is handled by distinct pipelines, with curated data undergoing minimal filtering and web data requiring extensive filtering and deduplication. This distinction guided the development of specialized algorithmic solutions for both pipelines. In addition to describing the processing methodologies, we provide an in-depth analysis of the datasets, increasing transparency and alignment with European data regulations. 
Finally, we share key insights and challenges faced during the project, offering recommendations for future endeavors in large-scale multilingual data preparation for LLMs. 
\end{abstract}
\section{Introduction}\label{sec.introduction}
In recent years, there has been an exponential increase in the development and deployment of 
large language models (LLMs) \cite{abdalla-etal-2023-elephant,abs-2405-14487}. The most 
prominent and accurate models are often the products of major technology companies and 
are available either through APIs (e.g. OpenAI's GPT models) or with non-open licenses 
(e.g. Meta's LLaMA family\footnote{Meta restricts commercial use based on monthly active users and restricts the use of LLaMa model to improve other LLMs.
}).
Moreover, these models are often US-centric, predominantly 
fluent in American English, and imbued with the values inherent to that cultural context.
Conversely, open and unrestricted models that focus on a variety of languages and 
cultures often emerge from academic settings. However, these models generally lack the 
accuracy and power of their more robust commercial counterparts. 
This disparity is primarily due to the differences in the resources available for 
training, such as the diversity and quality of data sources. The training data is 
crucial in shaping the behavior and values these models learn.
The OpenGPT-X\footnote{\url{https://opengpt-x.de/en/}} project aims to bridge this 
gap by merging the best of both worlds: delivering an open, unrestricted, high-performance 
model trained on diverse data that covers all major European languages. OpenGPT-X targets 
real-world use cases, e.g. from media and the automotive sector within the European Union.

In this paper, we present a comprehensive overview of the data processing pipelines developed 
for the OpenGPT-X project. Our focus is solely on the collection, preparation, and curation of the data, detailing the steps for data selection, conversion, normalization, quality filtering, and deduplication. The processed data was then made available for training of OpenGPT-X models \cite{opengpt_x_2024_13866365}, which utilized and sampled a subset of the data.
Our main contribution are:
\begin{itemize}

    \item A comprehensive outline of the data selection requirements, serving as practical guidelines for future projects.

    \item A detailed explanation of our multilingual data pipelines, highlighting distinct processing approaches for web and curated datasets.
    
    \item A thorough analysis of the processed datasets to ensure transparency and compliance with European data regulations (e.g. GDPR\footnote{General Data Protection Regulation, available at: \url{https://gdpr-info.eu/}}, Ethics Guidelines for Trustworthy AI\footnote{\url{https://digital-strategy.ec.europa.eu/en/library/ethics-guidelines-trustworthy-ai}}, Open Data Directive\footnote{\url{https://digital-strategy.ec.europa.eu/en/policies/legislation-open-data}}).
    
    \item Identification of project-specific challenges and lessons learned to guide future efforts in multilingual LLM development.
    
\end{itemize}

The remainder of this paper is organized as follows:  
Section \ref{sec:rel_work} provides context in the field of data processing and 
generation, specifically compiling a list of monolingual and multilingual datasets 
along with their methodologies.  
Next, the data selection process for curated data and web data is based on 
specific requirements that are outlined in Section \ref{sec:datasets}. This division leads to distinct software pipelines 
for processing each type in Section \ref{sec:pipelines}.
Results are analyzed separately for curated and web data in Section \ref{sec:analysis}, 
providing insights on their use for model training.
The paper also discusses challenges and lessons learned in Section \ref{sec:insights}, 
offering guidance on potential data, legal, and organizational obstacles in future projects.
Finally, we conclude the paper in Section \ref{sec:conclusion} and provide an outlook on future work.

\section{Related Work}
\label{sec:rel_work}
Training data for LLMs is primarily based on 
web-based data sources like Common Crawl\footnote{\url{https://commoncrawl.org/}}.  
In this section, we present an overview of similar initiatives 
to train LLMs to provide context for the data processing pipeline 
described in this paper. By outlining related datasets and their 
corresponding pipelines, we aim to highlight the key features 
and challenges of building such datasets, offering a comparative 
perspective to the approach taken in our 
work. We provide an overview in Table \ref{tab:datasets} (see Appendix).

\subsection{Monolingual datasets}
\label{sec:rel_work.mono}
The prevalence of English in LLM development is largely driven 
by the availability of English web data and the dominant role 
of English as a global language.
This abundance of English data allows better data quality control,
filtering, and deduplication by setting more aggressive selection criteria
and discarding larger amounts of suspicious data,
resulting in highly optimized datasets for LLM training.
However, this also led to a disproportionate focus on English, potentially limiting the applicability of models trained on these datasets to non-English contexts.

\textbf{Colossal Clean Crawled Corpus (C4)} \cite{raffel_shazeer_etal2020} 
is a widely-used dataset derived from CommonCrawl data and serves as the 
foundation for Google’s T5 model. C4 contains approximately 750GB of 
English text and is built through a filtering process designed to 
extract high-quality language data. The pipeline first filters out 
non-linguistic content, such as source code, placeholders, and documents 
with excessive punctuation or abnormal words per line ratios. A deduplication 
process is applied to spans of three sentences to ensure minimal repetition. 
Afterwards, documents filtered based on the score of language detection on
the textual content, keeping only English documents with a language score higher than 0.99, based
on the \textit{langdetect} tool\footnote{\url{https://github.com/Mimino666/langdetect}}. 
This document filter ensures that C4 remains highly focused on English content, with a 
restrictive approach to both content quality and language identification.

\textbf{The Pile} \cite{gao_biderman_etal2020} is a 800GB English dataset 
comprising a diverse set of 22 specialized text collections, making it 
one of the most comprehensive datasets for LLM training. It includes a 
wide variety of data sources, such as scientific papers, legal documents, 
fiction, patents, subtitles, and online forums, in addition to 
filtered CommonCrawl data. For the CommonCrawl portion, text is extracted 
using the jusText\footnote{\url{https://github.com/miso-belica/jusText}} 
text extractor, and language identification is performed using the \textit{pycld2} tool\footnote{\url{https://github.com/aboSamoor/pycld2}}. 
A \textit{fastText} classifier \cite{joulin_grave_etal2016b}, trained on 
the OpenWebText2 subset \cite{radford_wu_etal2019,gokaslan_cohen2019}, 
is used to further refine document quality. Deduplication is applied to both the 
CommonCrawl and OpenWebText2 subsets using the MinHashLSH algorithm \cite{broder1997}, 
ensuring minimal redundancy of the textual content while preserving its diversity. 
This pipeline design has made The Pile a key dataset for LLMs, especially due to 
its combination of web-sourced data with specialized collections.

\textbf{RefinedWeb} \cite{penedo_malartic_etal2023} is a 2.8TB English dataset 
created from filtered CommonCrawl data, employing the MacroData refinement 
pipeline. The pipeline begins by extracting text from WARC (Web ARChive) files using the
\textit{trafilatura} tool \cite{barbaresi2021}, which is then passed through 
several stages of filtration and deduplication.
First, the \textit{fastText} language classifier \cite{grave_bojanowski_etal2018}
identifies and retains English documents. The dataset undergoes both 
document-wise and line-wise heuristic filtering to remove low-quality 
or non-linguistic content. Deduplication is then applied in three stages: 
1) MinHashLSH deduplication at the document level, 
2) exact matching on spans of 50 tokens, and 
3) URL-based deduplication. 

\textbf{Dolma} \cite{soldaini_kinney_etal2024} contains about 11TB of English 
text and comprises of the CommonCrawl derived
data combined with code, social media, scientific papers, and fiction from 
the specialized web sites.
The CommonCrawl subset is extracted from the WARC files and is 
filtered by document-wise
and line-wise heuristics, the \textit{fastText} language classifier with 
the cutoff score of 0.5, and the \textit{fastText} toxic content classifier.
Deduplication is performed based on URLs, exact match of raw documents, and
on paragraphs using a Bloom filter \cite{bloom1970}.

\textbf{FineWeb} \cite{penedo_kydlicek_etal2024} contains approximately 43TB of 
English text and is constructed using 
DataTrove\footnote{\url{https://github.com/huggingface/datatrove}}, which 
was released alongside the dataset.
The pipeline builds upon the methodology used in RefinedWeb, beginning with text 
extraction from WARC files using the \textit{trafilatura} tool. 
Language filtering is performed with \textit{fastText}, applying a cutoff score 
of 0.65 to ensure the inclusion of high-quality English documents. After this, the 
dataset undergoes both document-wise and line-wise heuristic filtering to further
enhance quality by removing noise and non-linguistic content. The final deduplication 
step employs the \textit{MinHash} algorithm, ensuring a clean and diverse dataset that 
is well-suited for large-scale language model training.

\subsection{Multilingual datasets}
\label{sec:rel_work.multi}

Creating high-quality multilingual datasets presents distinct challenges compared to 
monolingual English datasets. Language identification becomes significantly more complex, 
as models must accurately differentiate between multiple
of languages, often with limited training data for lower-resource languages. Additionally, 
ensuring consistent quality across languages is difficult due to varying amounts of available 
web data, and some languages are overrepresented while others suffer from a lack of resources. 
High quality multilingual datasets are essential for training models that perform well across 
diverse linguistic and cultural contexts, enabling the training of LLMs that are 
not inherently biased toward dominant languages like English.

\textbf{CCNet} \cite{wenzek_lachaux_etal2019} is among the first datasets built on CommonCrawl data,
containing 3.2TB (1.5 billion documents) in 130 languages. The CCNet pipeline begins 
by extracting paragraphs from WET files and performing SHA-1-based deduplication. The 
extracted text is normalized by converting it to lowercase, replacing all numbers with 
zeros, and removing accents and punctuation.
Language detection is based on \textit{fastText}, with documents with a language score below 0.5  being 
discarded. The final step in the pipeline involves filtering documents based on 
perplexity scores calculated with language-specific 5-gram Kneser-Ney language 
models \cite{heafield2011}. The corresponding models are trained on tokenized Wikipedia 
data, using the sentence piece tokenizer \cite{kudo2018} to ensure consistent text segmentation.

\textbf{Multilingual C4 (mC4)} \cite{xue_constant_etal2021} extends the C4 dataset to 
101 languages and contains approximately 27TB of text. In contrast to C4, which focuses 
on English, mC4 replaces the language classification tool with 
\textit{cld3}\footnote{\url{https://github.com/google/cld3}} to handle the identification of multiple languages.
A significant modification in mC4 is the introduction of a filtering step that requires documents to contain at least three lines of text, each with 200 or more characters. This broader filter replaces C4’s original language-specific rule, which filtered based on English punctuation marks, ensuring a more consistent approach across languages and improving the quality of multilingual data.

\textbf{OSCAR 22.01} \cite{abadji_suarez_etal2022} contains about 8TB of text in 153 languages.
It is produced with the Ungoliant pipeline \cite{abadji_suarez_etal2021} from the CommonCrawl WET (Web Extracted Text) files.
The filtration starts with language detection, which utilizes
the \textit{fastText} model
to identify language of each line in a document.
If the confidence score is less then 0.8,
then the line is assigned with to the \textit{unknown language} class.
The proportion of each language in a document is calculated
as a percentage of bytes in the document assigned with that language.
A document is classified as multilingual 
if it contains at least 5 lines, not more than 5 languages, the
proportion of each language is at least ${1}/{(m + 1)}$ (where $m$
is the number of languages in the document) and the proportion of 
the unknown language is not larger than ${1}/{(m + 1)}$.
Otherwise, the document is classified as monolingual
and is assigned with the language having the highest weighted confidence
score. The weighted confidence of a language is calculated as a sum of products of
byte size and language confidence of each line classified to that language
divided by the total number of bytes in the document.
A monolingual document is passed to further processing steps
if the weighted confidence of its language is at least 0.6.
Language identification is followed by
the filtering of documents based on the line lengths, proportions
of characters of certain Unicode classes, and the URL based UT1 blocklist.\footnote{\url{https://dsi.ut-capitole.fr/blacklists/index_en.php}}
Deduplication is not applied to non-English data, and only line-wise deduplication
is applied to English data.

\textbf{BigScience ROOTS} \cite{laurenccon_saulnier_etal2022} is a dataset containing about 1.6TB of text in 46 natural languages.
This dataset comprises data from the crowdsourced list of 252 monolingual
and multilingual text collections, accompanied by documents extracted from the
CommonCrawl WARC files according to the list of domains suggested by the community
members. Extraction of text from HTML files is performed by custom code
inspired by the CommonCrawl extractor. Deduplication is performed on the level
of data sources. Additional data is obtained from the OSCAR version
21.09 dataset \cite{suarez_etal2020}. The OSCAR derived documents are passed through heuristic rules
based on word frequencies and are deduplicated in two steps using SimHash \cite{charikar2002,manku_jain_etal2007} and substring deduplication \cite{lee_ippolito_etal2022,manber_myers1993}.

The \textbf{MADLAD-400} \cite{kudugunta_caswell_etal2024} dataset contains about 30TB of text in 419 languages. The data
is obtained from the CommonCrawl dumps and is initially deduplicated on the line
level, followed by basic prefiltering similar to the C4 rules.
The semi-supervised language identification model
\cite{caswell_breiner_etal2020}, which is trained
with two tasks (supervised language detection task and
unsupervised corrupted input recovery task \cite{raffel_shazeer_etal2020}),
is used to classify the documents after that.
The language identification step is followed by another
set of quality filtering heuristics.
A few language-specific processing and filtering rules are applied based on the
manual inspection of 20 documents per language.


\textbf{RedPajama-v2}\footnote{\url{https://github.com/togethercomputer/RedPajama-Data}} 
contains about 20 billion documents and 30 trillion tokens
in 5 European languages.
It is produced with the extended CCNet pipeline and utilizes both heuristic based
and classifier based content filtering. The deduplication is performed on the document level using a Bloom filter.

The \textbf{HPLT} \cite{degilbert_nail_etal2024} dataset contains 50.1TB of text in 75 languages. The data is sourced from
the Internet Archive and CommonCrawl dumps in WARC format. First, text is
extracted using the \textit{warc2text} tool\footnote{\url{https://github.com/bitextor/warc2text}} and classified by language
using the CLD2 library\footnote{\url{https://github.com/CLD2Owners/cld2}}. Next, text is cleaned from encoding errors,
and another two-stage paragraph-level LID is performed. Its first stage
utilizes the fastText model, and the second stage applies dictionary based
spell checking for several languages related to the fastText result and
selects the language, for which fewer mistakes are detected 
\cite{banon_ramirez_etal2024}. Finally, MinHash based deduplication is 
performed on the document level.

The collection of data we processed comprises data from both web and curated sources,
including parts of The Pile and RedPajama-v2, and the content and metadata 
are normalized in a uniform way
by the data processing pipeline described in this work.



\section{Data selection}
\label{sec:datasets}

The performance of large language models in downstream tasks benefits from large volumes of diverse, high-quality training data. Key properties for effective pretraining include the diversity of knowledge, domains, and tasks, which enhance a model's generalization abilities \cite{gao_biderman_etal2020}, as well as the appropriate size to fully saturate the model's learning capacity \cite{jordan_hoffmann_etal2022}. To mitigate cultural bias from monolingual English datasets, we emphasize the selection of non-English data. Specifically, we aim to balance our data to encompass all European languages. 
This section outlines the data selection requirements that ensure these desirable properties and provides an overview of the chosen data sources.

Given the aforementioned properties, we suggest a method for data selection that ensures a large volume\footnote{The amount of training data is typically measured by the number of tokens or words. However, the choice of a tokenizer affects the number of generated tokens, making absolute numbers for different models not directly comparable. Therefore, we use the \texttt{wc} command line utility to estimate the number of words.} of diverse and \textsl{high-quality} after preprocessing.

\textbf{Data Quantity and Diversity.}
A significant portion of our data is derived from crawled web pages, complemented with curated datasets to ensure a broad spectrum of languages, genres, text types, and domains. These curated datasets are carefully selected to enhance performance in various downstream tasks such as question answering, machine translation, and summarization.

\textbf{Data Quality.} 
The quality of large language models is highly dependent on the textual training data, often measured in terms of the number of tokens or words. High-quality data should be multilingual (covering as many relevant languages as possible), diverse (sourced from multiple domains and document types), free of toxic or offensive content (to prevent harmful outputs), and unbiased (to ensure fair model behavior). In this context, data quality directly reflects the properties of the text itself. For a detailed discussion of our filtering processes, see Section  \ref{sec:pipelines}.

\subsection{Selected data}
\label{sec:datasets.selection}

Following these requirements, our data sources can be categorized into two groups. The first group consists of curated data from multiple existing datasets, providing a relatively small but diverse set of high-quality knowledge sources. The second group comprises general web data sourced from the CommonCrawl project. 

\subsubsection{Curated data}
\label{sec:datasets.selection.curated}

Curated datasets are essential for providing high-quality, diverse training data for LLMs. These datasets have typically undergone a quality review process, either prior to their release or through ongoing public review,
and are publicly available
(see Appendix \ref{sec:appendix.curated} for a list of curated datasets we compiled). The list includes data such as Wikipedia, collections of 
scientific articles, books, and source code.

To select appropriate datasets, we employed the following considerations:

\begin{enumerate}
    \item Legal and Licensing
    \begin{itemize}
        \item Is the license unproblematic? (e.g., allows research and commercial use, is permissive, and does not include restrictive copyleft clauses). 
        \item Is the data GDPR-compliant?
    \end{itemize}
    
    \item Language and Relevance
    
    \begin{itemize}
        \item Does the dataset contain documents in a language relevant to our research?
        \item Does the dataset contain a sufficient quantity of tokens/words/documents? 
        \item Do the documents cover relevant topics? 
        (e.g. for downstream applications)
    \end{itemize}
    
    \item Quality and Integrity
    
    \begin{itemize}
        \item Is the content of documents error free? 
        (e.g., no artifacts from Optical Character Recognition, misspellings, or antiquated language)
        \item Are documents from a geographical region of interest? 
        \item Are the documents from a relevant time period?
        (e.g. no historic documents)
        \item Are the documents human-generated\footnote{While it is not always easy to detect AI-generated content, extra caution should be applied for datasets created after 2021 to avoid including synthetic data.}?
        \item Is the dataset artificial?
        (e.g. no automatically generated content or synthetic data)
        \item Is there no significant overlap with other datasets already in use?
    \end{itemize}
    
    \item Resource Availability
    
    \begin{itemize}
        \item Are there sufficient resources available to preprocess the data? 
        (e.g., adequate disk space and computational resources)
        \item Has the dataset not been fully or partially superseded by a newer version?
        (e.g. no outdated version)
    \end{itemize}

\end{enumerate}

If one or more of these questions were answered negatively, the dataset was excluded from further processing. These criteria ensure that the curated datasets selected for our project are of the highest quality, relevance, and suited for our specific objectives.

\subsubsection{CommonCrawl data}
\label{sec:datasets.selection.cc}

Unlike curated data, web data is available in a large amount. For this reason, we prioritize processing web data only based on recency and language. Specifically, we use data from CommonCrawl (CC), an organization that maintains a free, open-source repository of web data accessible to anyone since they provide the largest amount of data. They perform monthly crawls of internet webpages, compiling them into data dumps labeled in the format YYYY-WW (e.g., 2024-22, see Section \ref{sec:analysis.web} for a list of selected dumps). 
CommonCrawl dumps can be downloaded from URLs\footnote{Following the schema \texttt{https://data.CommonCrawl.org/crawl-data/CC-MAIN-<YEAR-MONTH>/index.html}.} in two main formats: WARC and WET. WARC (Web ARChive format) files contain raw data from the crawl, including the full page HTML and request metadata. WET (WARC Encapsulated Text) files provide a text-only version of the websites, excluding the web metadata.
In our pipeline, we use the WET files exclusively as they provide the extracted textual content of web pages in a simplified format. This eliminates the need for extensive parsing of raw HTML data found in WARC files, allowing us to focus directly on text analysis. For instance, the CC dump ``2024-22'' includes data crawled from 2.5 billion web pages, totaling 18.35 TiB of compressed HTML data. This particular dump consists of 90,000 WET files, which are grouped into segments for processing.

\section{Data pipelines}
\label{sec:pipelines}

This section describes the data pipelines we used for processing both curated and web data. Given the distinct nature and processing requirements of these two types of data, the pipelines are described in separate subsections.

\subsection{Pipeline for curated data}
\label{sec:pipelines.curated}

\begin{figure}[ht]
    \centering
    \includegraphics[width=\linewidth]{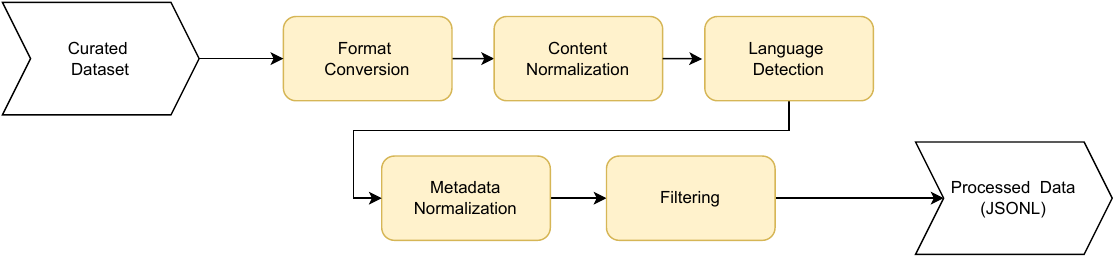}
    \caption{\label{fig:curated_pipeline}%
    Data pipeline for curated datasets: The pipeline involves format conversion, content normalization, followed by language detection, content normalization and filtering. The output is high-quality preprocessed data in JSONL format.}
\end{figure}

Curated datasets are published in a variety of formats and metadata. In contrast to web crawled data, these datasets consist of thematically related and high-quality texts, due to the previous curation process (e.g. reviewing for scientific articles). Therefore, a quality-based filter is less necessary and the preprocessing is based the pipeline in Figure \ref{fig:curated_pipeline}.

\textbf{Format Conversion.}
The first step in processing curated data involves converting the raw data into JSONL format. Depending on the raw data's format, we employ different processing tools. For PDF documents, \textit{Grobid}\footnote{\url{https://github.com/kermitt2/grobid}} converts PDFs to XML/TEI as an intermediate format which we then convert to JSONL to identify and remove layout elements from the content. 
For Wikimedia datasets (e.g., Wikipedia, Wikibooks), \textit{wikiConverter} parses the Wikimedia XML dumps and converts them to JSONL. The data is first converted using \textit{mwparserfromhell}\footnote{\url{https://github.com/earwig/mwparserfromhell}}, then list pages and disambiguation pages are removed, and finally the remaining data is processed as with other curated datasets. 
For any other format (e.g. plain text, csv, sql dumps, etc), we created custom conversion tools.

\subsubsection{Normalization}
\label{sec:pipelines.curated.normalization}
Normalizing the content is important to ensure that the rest of the pipeline consistently handles data, facilitating versioning, processing, and sharing. We implement two types of normalization: content normalization and metadata normalization.

\textbf{Content Normalization.}
Content normalization ensures that the text data is uniformly formatted and encoded. 
(i) All text data is encoded in UTF-8 to maintain a standard character encoding format.
(ii) We apply NFKC normalization\footnote{Normalization Form KC as described in \url{https://unicode.org/reports/tr15/##Norm_Forms}} to the text content, converting characters to their canonical decomposition followed by compatibility composition to ensure that visually similar characters are represented consistently.  
(iii) Excessive white space is trimmed, and inconsistent spacing is corrected. 
(iv) Finally, we remove common conversion artefacts such as HTML tags or special characters to ensure the text is free from noise.

\textbf{Metadata Normalization.}
Metadata normalization is important to maintain a consistent set of essential metadata attributes that provide a snapshot of each document. This metadata forms the basis for subsequent analysis and filtering steps.
The format for a single document (a line in a JSONL file) is as follows:

\begin{verbatim}
{
    "meta": {
        "docid": <corpus/language/fileno/docno>,
        "url": <url:String>,
        "title": <title:String>,      
        "download_date": <ISO-date:String>",
        "language": <ISO-language-code:String>,
        "language_score": <language-detection-score:Float>},
    "text": <text:String>
}
\end{verbatim}
\noindent
Where
\begin{itemize}
\item \texttt{meta} contains all document metadata.
\item \texttt{docid} is a unique ID comprised of the corpus name, the number of the file a document originates from, and the running document number within this file.
\item \texttt{url} is a string representing the URL of a document (where appropriate).
\item \texttt{title} is the title of a document (where appropriate).
\item \texttt{download\_date} is the download date in YYYY-MM-DD format.
\item \texttt{language} is a 2- or 3-letter ISO language code \cite{iso639} for the language of the content.
\item \texttt{language\_score} is a score estimated from language detection \cite{joulin_grave_etal2016a} (1.0 where the language is not ambiguous); ``xx'' is used for non-language content such as source code.
\item \texttt{text} is the normalized document content. 
\end{itemize}
For an example document, see Appendix \ref{ssec:appendix.metadataexample}.
Note that the example for the document format 
already includes information from language detection.

\subsubsection{Language Detection}
\label{sec:pipelines.curated.langdetect}

Following content normalization, language identification is performed using \textit{fasttext} \cite{joulin_grave_etal2016a}. The detected language code is included both in the metadata field and as part of the filename for the pipeline output to enable grouping of data by language. 

\subsubsection{Filtering}
\label{sssec:pipelines.curated.prefiltering}

To ensure appropriate data quality, we aim to remove low-quality documents as early as possible in the processing pipeline. 

We apply a set of filters to remove some outliers:
\begin{itemize}
    \item Documents with a language score lower than 0.5 are filtered out.
    \item Documents that contain fewer than 200 characters are removed.
\end{itemize}

These thresholds (200 characters, 0.5 language score) were chosen based on settings used in previous projects (e.g., HPLT\footnote{\url{https://hplt-project.org/datasets/v1}}) and are designed to ensure that only relevant and high-quality data proceeds to the subsequent stages of processing.

The final output of the curated data pipeline is a set of JSONL files, organized by language and corpus, ready for further analysis and model training.

\subsection{Pipeline for web data}
\label{sec:pipelines.web}

\begin{figure}[ht]
    \centering
    \includegraphics[width=\linewidth]{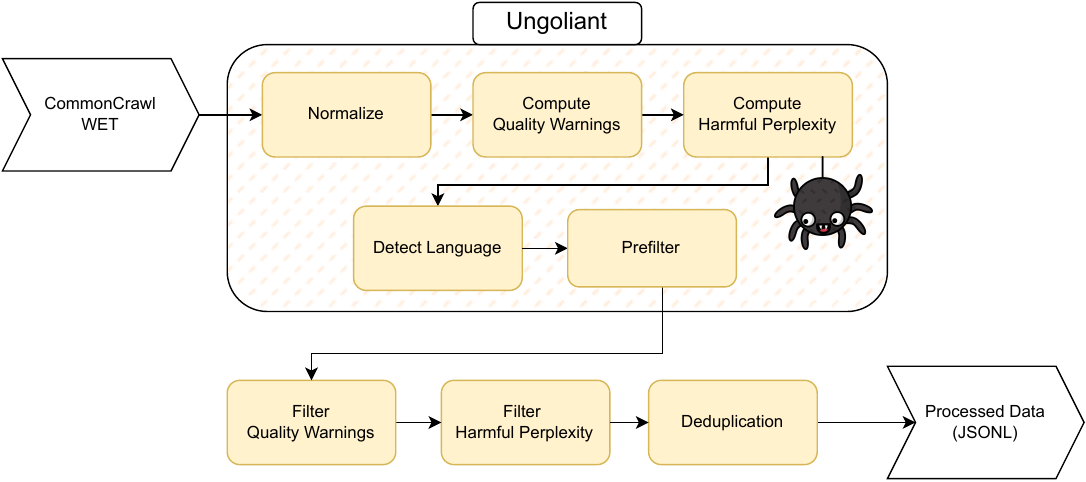}
    \caption{\label{fig:web_data_pipeline}%
    Overview of the Web Data Processing Pipeline: The pipeline processes WET files with Ungoliant by normalizing, computing quality warnings and harmful-perplexity, detecting language, and prefiltering. Afterwards, data is filtered based on quality and harmful-perplexity before deduplication, producing high-quality preprocessed data in JSONL format.
   }
\end{figure}

Processing web data from large-scale sources like CommonCrawl presents significant challenges due to its unstructured and noisy nature. To transform the raw web data into a suitable format for training LLMs, we utilized a specialized pipeline (see Figure \ref{fig:web_data_pipeline}). This section provides a detailed breakdown of each processing stage, including normalization, filtering, and deduplication, using the Ungoliant pipeline.

\subsubsection{Dataset Acquisition and Preparation}
\label{sec:pipelines.web.prep}

The process begins by downloading CommonCrawl dumps in WET file format using a custom download script. These dumps contain the textual content and metadata of web pages, compressed in \texttt{.txt.gzip} format. After retrieving the raw data, we organized it into separate folders based on the dump year for easier processing.

\subsubsection{Ungoliant Pipeline}
\label{sec:pipelines.web.ung}

The core of the web data processing relies on the Ungoliant pipeline \cite{abadji_suarez_etal2021}, a modular system optimized for handling CommonCrawl corpora and producing an OSCAR-like dataset \cite{abadji_suarez_etal2022}. The Ungoliant pipeline is conceptually split into several key components:

\begin{itemize}
    \item \textbf{Normalization}: Similar to the normalization applied to curated data (see Section \ref{sec:pipelines.curated}), this step ensures consistency in text encoding, removing noise, normalizing text formatting, and encoding all content into UTF-8.
    \item \textbf{Computation of quality warnings}: Ungoliant generates quality warnings for each document which are then used for subsequent filtering stages.
    \item \textbf{Computation of harmful-perplexity}: Harmful content is identified using perplexity scores based on a pretrained KenLM model \cite{jansen_tong_etal2022}. This model evaluates documents to determine whether they contain harmful content (e.g., adult material).
    \item \textbf{Language detection}: Sentence-based language identification is performed using embedded\footnote{Ungoliant supports the detection of 176 languages. } pretrained fastText models \cite{joulin_grave_etal2016a,joulin_grave_etal2016b}. 
    \item \textbf{Prefiltering}: Documents are filtered based on criteria similar to those used for curated data (see Section \ref{sssec:pipelines.curated.prefiltering}), such as removing documents with a low number of characters or low language detection scores.
\end{itemize}

\subsubsection{Filtering}
\label{sec:pipelines.web.filtering}

After the documents pass through the Ungoliant pipeline, we apply additional filtering steps based on quality warnings and harmful-perplexity scores to further refine the dataset.

\textbf{Quality Warning filtering.} Ungoliant flags documents with several quality warnings: \texttt{tiny}, \texttt{noisy}, \texttt{header}, \texttt{footer}, and \texttt{short\_sentences}. These warnings are accompanied by predefined thresholds, and documents that exceed these thresholds are filtered out. The thresholds are defined as follows:
 \begin{itemize}
     \item \texttt{tiny}: When the document contains fewer than 5 sentences (lines).
     \item \texttt{noisy}:  When the ratio of non-letters to the total number of characters exceeds 50\%.
     \item \texttt{header}: Defined through a three-step process:
     \begin{itemize}
        \item Iterate through the first 20\% of a document\footnote{The percentage is calculated on a line-base,  i.e. when the character '$\backslash n$' is found.}. 
         \item Identify short sentences (lines) with fewer than 100 characters.
         \item Annotate the document as \texttt{header} if more than 50\% of the sentences in this section are short.

     \end{itemize}
     \item \texttt{footer}: Similar to the header but applied in reverse order to the above one.

     \item \texttt{short\_sentences}: When the document has a high number ($\ge 50\%$) of lines with less than 100 characters. 
 \end{itemize}

\textbf{Harmful-Perplexity Filtering.}
Filtering adult content is a critical part of ensuring that the models are trained on safe and appropriate data. We use the perplexity scores provided by the KenLM-based model trained on adult content from the UT1 Blacklist \cite{jansen_tong_etal2022} as part of the Ungoliant pipeline. Lower perplexity scores indicate a higher likelihood that the document contains harmful or adult content. We set a perplexity threshold of 5, meaning any document with a perplexity score below this threshold is filtered out.

\subsubsection{Document deduplication}
\label{sec:pipelines.web.dedup}

Building on the quality filtering performed in the previous stage, the next step after running the Ungoliant pipeline is document deduplication. Deduplication refines the dataset by removing both exact and near duplicate documents, thus reducing the overall size of the data.
After extensive testing \cite{leveling_helmer_etal2024}, we found that the MinHash/LSH algorithm offers the best balance of precision, recall, and computational efficiency for deduplicating large volumes of web data. The deduplication process was executed on a per-dump and per-language basis (``local'' deduplication), which is both convenient and backed by newer research \cite{penedo_kydlicek_etal2024} that confirms that \emph{local} deduplication is more favorable than \emph{global} (across dump and/or language) approaches for preserving data diversity while minimizing redundancy.

\subsection{Technical Infrastructure}
\label{sec:pipelines.infrastructure}

To  effectively manage the extensive data processing requirements, we leveraged significant computational resources. The processing of large-scale datasets, including both curated datasets and CommonCrawl data, required the use of high-performance computing infrastructure.

Initially, a NVIDIA DGX-2 machine was employed for downloading and processing the curated datasets, as well as for deduplicating the first CommonCrawl dumps. As the volume of data increased, the processing of the remaining CommonCrawl datasets was distributed across two large compute clusters.

The technical specifications of these compute clusters are as follows:

\begin{itemize}
    \item Compute Cluster 1: Provided by TU Dresden, consisting of compute nodes with 2 x Intel Xeon Platinum 8470 (52 cores) @ 2.00 GHz CPU and 512 GB RAM.
    \item Compute Cluster 2: Provided by FZ Jülich, consisting of compute nodes with 2 x Intel Xeon Platinum 8168 (2x24 cores) @ 2.7 GHz CPU and 192 GB RAM.
\end{itemize}

These compute resources enabled us to handle the extensive data processing requirements necessary for pretraining our large language models.

\section{Data Analysis}
\label{sec:analysis}

This section provides a detailed analysis of the datasets used in our project, focusing on both curated and web data sources. The goal is to offer a comprehensive overview of the data characteristics.

We begin by presenting statistics for the curated datasets (Section \ref{sec:analysis.curated}), including details on corpus names, languages, text types, and other relevant metadata. We then analyze the web data (Section \ref{sec:analysis.web}), examining the size, composition, and the effects of filtering and deduplication. This includes a breakdown of the number of documents, word counts, and language distributions.

It is important to note how LLM training often uses sampling techniques to optimize data distribution, considering factors like language balance and the fertility score \cite{ali_mehdi_etal_2024}. 
In this paper we focus on techniques to process raw data into high-quality data for subsequent steps in LLM training, e.g. sampling. 

\subsection{Curated Data}
\label{sec:analysis.curated}

As previously mentioned, curated data undergoes a distinct processing pipeline. This is due to the assumption that curated datasets generally maintain a higher level of quality, thereby requiring less intensive filtering and deduplication compared to web data. A comprehensive list of the curated datasets used in our study is provided in Table \ref{tab:curated_data_list}, which details 75 datasets, including information on language, format, license, domain, and the number of documents/words, along with the filtering percentage. In the subsequent sections, we will conduct a detailed analysis of the most significant columns.

\begin{figure}
    \centering
    \includegraphics[width=0.6\linewidth]{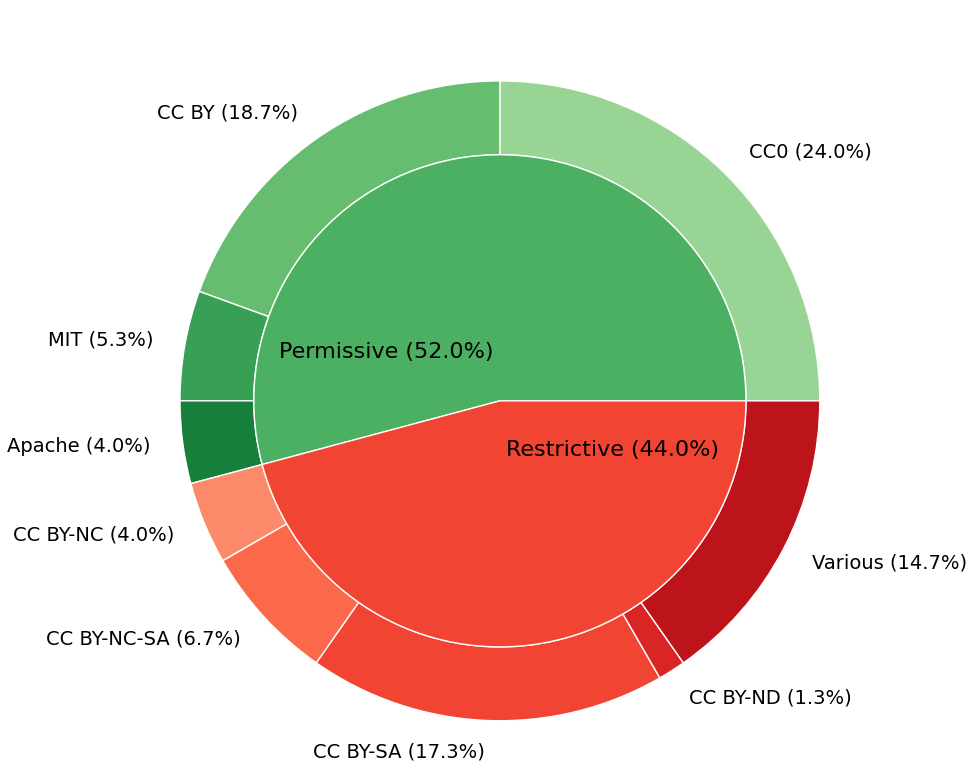}
    \caption{\label{fig:licenses}%
    Distribution of curated dataset licenses. Most datasets have permissive licenses (CC0, CC BY, MIT, Apache), 
    while restrictive licenses are primarily represented by datasets sourced from multiple origins (\textit{Various}). }
    
\end{figure}
\subsubsection{Licenses} 
As outlined in the data selection process (see Section \ref{sec:datasets.selection.curated}), selecting datasets according to legal and licensing constraints is a crucial first step. Figure \ref{fig:licenses} illustrates the distribution of licenses across the curated
datasets used in this project. The majority (52\%) of the selected datasets' licenses are permissive and not restricted for commercial use, featuring licenses like CC0 (24\%), CC BY (18.7\%), MIT (5.3\%), and Apache (4\%)\footnote{To simplify aggregation, the license versions are omitted here and in Figure \ref{fig:licenses}. However, in Table \ref{tab:curated_data_list}, the specific version numbers for all the datasets are clearly indicated.}. These licenses enable broad usage of the data, including for commercial purposes, and promote ease of dissemination and integration in various projects.

Conversely, 44\% of the datasets are categorized under restrictive licenses, which limit their use, especially for commercial applications. These include licenses such as CC BY-NC-SA (6.7\%), CC BY-NC (4\%), and the ``Various'' category (14.7\%), which represents datasets compiled from multiple sources and classified under the most restrictive license to mitigate legal risks.

An important observation is that many datasets did not explicitly mention a license on their official website. For these datasets, we conducted further investigation into the sources, often uncovering related licensing information. However, for some datasets (marked with a dagger in Table \ref{tab:curated_data_list}), we were unable to directly determine the correct licenses despite extensive efforts. 

\subsubsection{Languages}
\begin{figure}
\centering
\begin{subfigure}{.5\textwidth}
  \centering
  \includegraphics[width=0.95\linewidth]{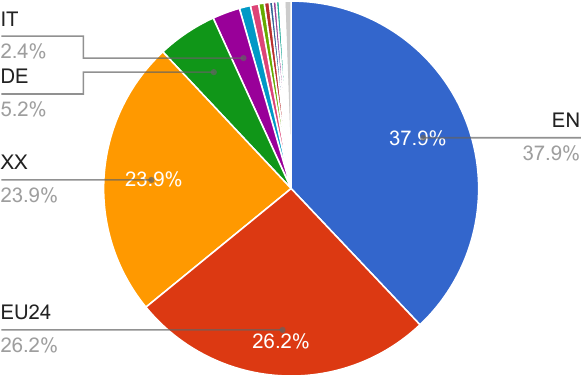}
  \caption{Complete word distribution} 
  \label{fig:curated_lang_word_a}
\end{subfigure}%
\begin{subfigure}{.5\textwidth}
  \centering
  \includegraphics[width=0.95\linewidth]{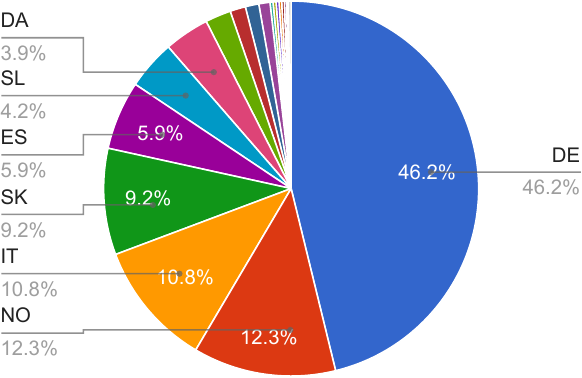}
\caption{Word distribution excluding EN, EU24, and XX.}
\label{fig:curated_lang_word_b}
\end{subfigure}
 \caption{Word distribution across different languages in the curated datasets.}
\label{fig:curated_lang_word}
\end{figure}

Our curated dataset encompasses 24 unique languages. For clarity, datasets that include a substantial portion of the official 24 EU languages are grouped under the category “EU24”. Additionally, the analysis also features statistics for other European languages, such as Norwegian (NO). The label “XX” is used to indicate datasets focused on source code.

As shown in Figure \ref{fig:curated_lang_word_a}, English (EN) accounts for 37.9\% of the total word count, followed by the EU24 languages at 26.2\%, and source code (XX) at 23.9\%. Together, these three categories make up 88\% of the curated data. The inclusion of source code is particularly important in this context, as it provides curated, high-quality examples for training models on coding tasks.

Since many datasets feature multiple languages, specific per-language statistics are not always provided, but rather an overall distribution is emphasized.  On average, multilingual datasets contain significantly fewer words (712.8 million) compared to monolingual (4.38 billion) ones. Indeed,  while contributing to language diversity, they often lack the volume of content found in larger monolingual datasets.

After removing the dominant categories (EN, EU24, and XX), Figure \ref{fig:curated_lang_word_b} reveals that German (DE) constitutes 46.2\% of the remaining data. This significant proportion reflects the regional focus of the project, as it is based in Germany, and emphasizes the role of localized datasets in shaping the overall corpus. Other notable languages include Norwegian, Italian, and Spanish, though their contributions remain relatively smaller compared to German.


\paragraph{Analysis of Word Distribution and Dataset Representation in Multilingual Corpora}
One would expect the word distribution to follow the number of datasets available; however, correlation analysis reveals this is only partially true. 
The Pearson correlation indicates a stronger linear relationship (0.855, $P < 0.001$). 
By using linear regression, we observe notable deviations in certain languages. For instance, German (DE) exhibits a significant negative residual, with the actual word count being 5.1 billion fewer than expected, despite having a high number of diverse datasets. This suggests that while German datasets are numerous, they tend to be smaller in size or less word-dense compared to other languages. On the other hand, English (EN) has a positive residual, showing 3.88 billion more words than predicted, which can be attributed to the oversampling of English datasets in the corpus. In contrast, French (FR) underperforms relative to its dataset count, with a negative residual of 1.54 billion, indicating that the French data is underrepresented in terms of word volume.

This analysis underscores the need for balanced sampling across languages. While oversampling English may provide more content for model training, undersampling key languages like French could result in biases that limit the multilingual capabilities of the models. Hence, we recommend strategic oversampling of underrepresented languages and careful moderation of overrepresented languages to ensure linguistic diversity and fairness in the curated dataset.

\subsubsection{Domains}

\begin{table}
\centering
\begin{tabular}{lrrrr}
\hline
Domain                 & \# Datasets & Avg. Words/DS [M] & Total Words [M] & Percentage [\%] \\ \hline
Source Code            & 1        & 73.064                & 73.064      & 40.46           \\
Law and Administration & 22       & 1.674                 & 36.839      & 20.40           \\
Web                    & 6        & 4.120                 & 24.721      & 13.69           \\
Medical                & 5        & 3.229                 & 16.147      & 8.94            \\
Math                   & 6        & 2.684                 & 16.105      & 8.92            \\
Forum                  & 2        & 3.655                 & 7.311       & 4.05            \\
Books                  & 5        & 752                   & 3.763       & 2.08            \\
News                   & 4        & 473                   & 1.893       & 1.05            \\
Knowledge Base         & 4        & 105                   & 423         & 0.23            \\
Culture                & 3        & 48                    & 146         & 0.08            \\
Recreation             & 2        & 73                    & 146         & 0.08            \\ \hline
\end{tabular}
\caption{Total words, average words per dataset (Avg Words/DS), and documents for each domain, sorted by total words. }
\label{tab:word_distribution_per_domain}
\end{table}

The curated dataset spans multiple domains, each contributing a different share to the total word count. Table \ref{tab:word_distribution_per_domain} provides a breakdown of the number of datasets, the average word count per dataset, total word count, and the percentage contribution of each domain.

Source code emerges as the largest domain by word count, accounting for 40.46\% of the total curated data, with over 73 billion words. This is followed by law and administration, which contributes 20.40\%, and the web domain at 13.69\%. Collectively, these three domains represent 74.55\% of the total word count. This strong presence of technical, legal, and digital content suggests that the curated dataset is well-suited for training models focused on tasks related to programming, legal reasoning, and web-based applications.

On the other hand, smaller domains such as culture (0.08\%), recreation (0.08\%), and knowledge base (0.23\%) contribute much less to the overall dataset. These domains are likely underrepresented either due to the limited availability of datasets or their inherently smaller size.

\subsubsection{Sizes}

\begin{figure}
    \centering
    \includegraphics[width=0.7\linewidth]{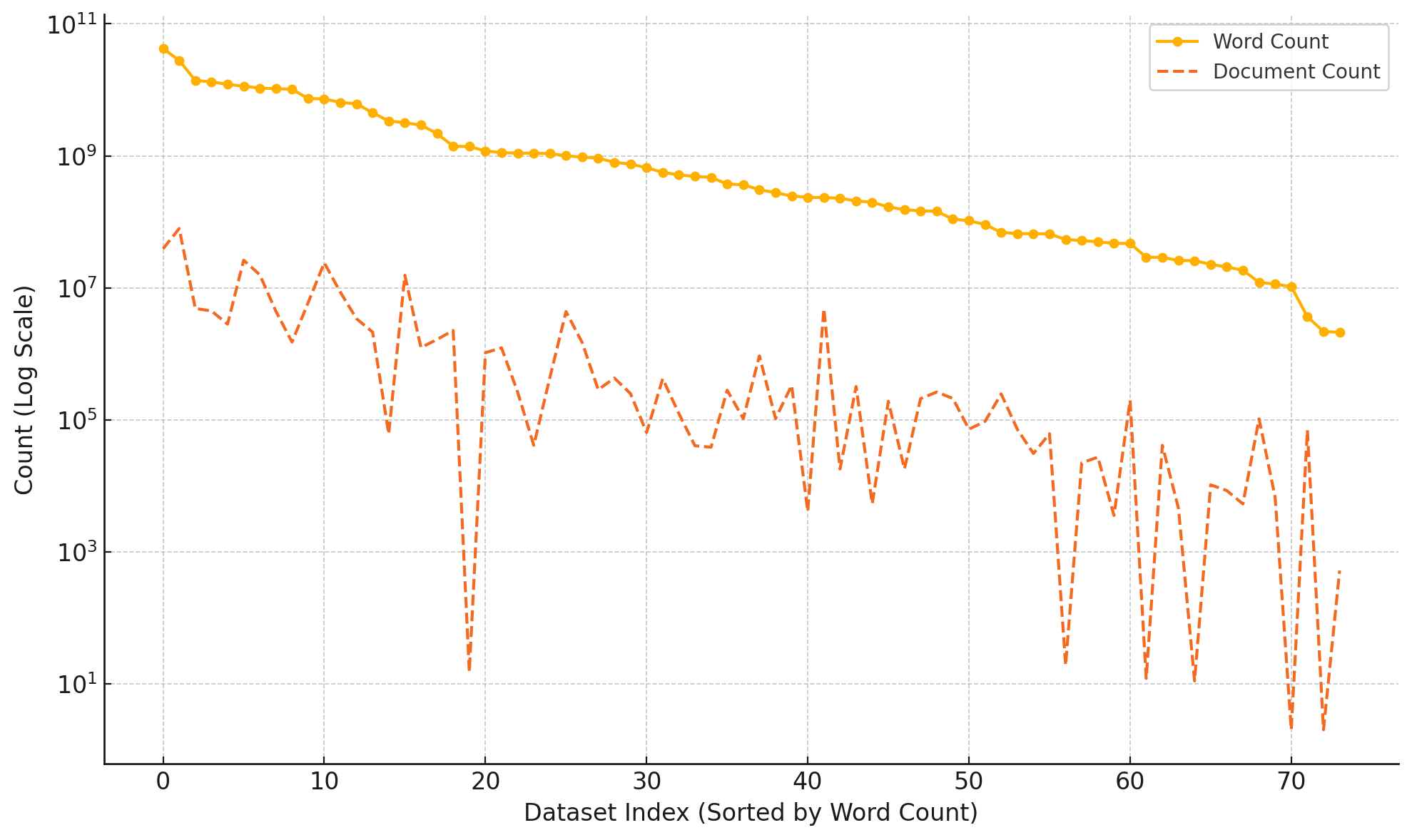}
 \caption{Log-scaled comparison of word counts and document counts across datasets, sorted by word size. Some datasets exhibit a notably high word count despite having relatively few documents, indicating longer than average document sizes.}    \label{fig:curated_size}
\end{figure}

The distribution of word counts within the curated dataset is highly skewed, with a few large datasets contributing the majority of the total word volume as shown in Figure \ref{fig:curated_size}. Initially, the largest contributors were \textit{StarCoder}, \textit{EurLex}, and \textit{MaCoCu} , together accounting for 50\% of the total word count. However, since \textit{StarCoder}  focuses on source code rather than natural language, it was excluded from further analysis, and the word count distribution was recalculated.

In the revised analysis, four datasets (\textit{peS2o}, \textit{MaCoCu}, \textit{Legal MC4},  and \textit{EurLex})  now make up 50\% of the total word count. Expanding this to 70\%, the dataset contributions broaden to include three additional Pile subsets (\textit{PMC extracts}, \textit{Openwebtext2}, and \textit{Free Law Opinions V2}), as well as \textit{Wikimedia Wikipedia}. Notably, 17 datasets account for 90\% of the total word count, underscoring the significant concentration of data within a limited number of large datasets.

Additionally, certain datasets display an unusually high word count relative to their document counts. A regression analysis was performed to better understand these discrepancies, calculating residuals to detect significant deviations from the expected document-to-word ratio. This analysis revealed the following outliers:

\begin{itemize} 
\item \textit{Spanish legal corpora (ES)}: This legal corpus consists of only 15 documents but contains over 1.38 billion words. The residual analysis (5.4) shows that the word count is approximately 221 times higher than expected based on the number of documents. This suggests that individual documents within this corpus are unusually large.

\item \textit{Projekt Gutenberg (EU24)}: A well-known collection of books, \textit{Projekt Gutenberg} contributes over 3.37 billion words with only 60,912 documents. This results in a positive residual of 1.8, indicating that the dataset's word count per document is significantly higher than the regression model predicted. This observation aligns with expectations for book collections, as books typically contain more words per document compared to other formats such as articles or reports.

\item \textit{Pile: PMC extracts (EN)}: A medical dataset with 2.8 million documents and over 12.1 billion words, \textit{PMC extracts} exhibits a positive residual of 1.07. This indicates that the word count is significantly higher than predicted by the model. This is expected, as the dataset primarily contains full-length medical research articles, which are generally more detailed and content-rich compared to other document types.

\end{itemize}

\subsubsection{Filtering}

Across the curated datasets, the average filtering percentage is 5.33\% ($\pm 5.98\%$). Based on these values, filtering can be  split in three categories:

\begin{itemize}
    \item Low Filtering ($<1\%$): Datasets that underwent minimal filtering, typically because they contained well-structured, clean data to begin with. 
    
    \item  Medium Filtering (1\% to 15\%): The majority of datasets fall into this category, with moderate filtering applied to remove noise or short documents without heavily impacting the total word count. 
    
    \item  High Filtering (>15\%): Datasets with a high filtering percentage, indicating substantial noise or irrelevant content. 
\end{itemize}

In reviewing the impact of filtering, we found that documents with low language scores typically fall into one of four categories: i) mixed-language documents, ii) documents with conversion errors containing numerous special characters, iii) documents involving closely related languages (e.g., Croatian and Serbo-Croatian), and iv) documents in low-resourced languages where the content was misidentified as a different language. 

To further understand the impact of filtering across datasets, we explored potential correlations between the percentage of filtered data and several key dataset characteristics: the number of words, number of documents, format, language, and domain. Our analysis showed only modest correlations in most cases, indicating that filtering tends to act somewhat independently of these factors (refer to Appendix \ref{sec:appendix.curated.filtering} for a more detailed analysis).

\subsubsection{Summary}
The curated data presents a comprehensive and diverse collection of 75 datasets spanning 25 languages, various domains, and multiple formats, offering a robust foundation for a wide range of research tasks. 
The variety within the data makes it a valuable asset even though there are imbalances in the distributions originating from large datasets dominating the total word count.  These imbalances can be remedied by subsequent sampling techniques to balance the data. The prevalence of technical and legal data, alongside source code, reflects the strengths of the preprocessed data in supporting tasks in these fields.

While filtering in general is necessary to ensure data quality, its impact on the curated datasets is relatively moderate, with only a few datasets requiring significant cleaning. The weak correlations found between filtering percentages and dataset characteristics suggest that filtering is driven more by the specificity of the dataset than by its size, format, or domain. 
We found that a strategic approach to data selection is a key challenge. Leveraging the wide domain and language diversity can help ensure more balanced outcomes. 



\subsection{Web Data}
\label{sec:analysis.web}

As previously mentioned (Section \ref{sec:datasets.selection.cc}), the web data used in this project was sourced from CommonCrawl. For this study, we utilized 60 distinct dumps from CommonCrawl, spanning a broad timeframe. The earliest dump was week 42 of 2014 (2014-42), and the most recent dump was week 5 of 2023 (2023-5). This extensive range allowed us to capture a wide variety of web content across nearly a decade, ensuring that the dataset reflects both historical and contemporary web information.

On average, each CommonCrawl dump contains approximately 2.7 billion documents. Across all 60 dumps, we accumulated a total of 173 billion documents, representing around 703 terabytes of raw, unprocessed data. This vast volume of data provided a strong foundation for training large-scale language models but also introduced significant challenges in terms of data processing and filtering, as discussed in previous sections.

\begin{table}[htb!]
\centering
\begin{tabular}{cl}
\toprule
\textbf{Year} & \textbf{Week} \\ 
\midrule
2014         & 42               \\ 
2015         & 14, 48           \\ 
2016         & 22, 44           \\ 
2017         & 13, 47, 51       \\ 
2018         & 5, 9, 13, 17, 22, 26, 30, 34, 39, 43, 47, 51 \\ 
2019         & 4, 9, 13, 18, 22, 26, 30, 35, 39, 47, 51 \\ 
2020         & 5, 10, 16, 24, 29, 34, 40, 45, 50 \\ 
2021         & 4, 10, 17, 21, 25, 31, 39, 43, 49 \\ 
2022         & 5, 21, 27, 33, 40, 49 \\ 
2023         & 6, 14, 23, 40, 50 \\ 
\bottomrule
\end{tabular}
\caption{List of dumps by year and week.}
\label{tab:dumps_by_year}
\end{table}

\paragraph{Dumps}
In this section, we provide a detailed look at the CommonCrawl dumps used in our project. The dumps span multiple years, with varying distributions of weeks per year. Table \ref{tab:dumps_by_year} presents  the list of years and the corresponding weeks for each dump.



\begin{figure}
\centering
  \centering
  \includegraphics[width=0.75\linewidth]{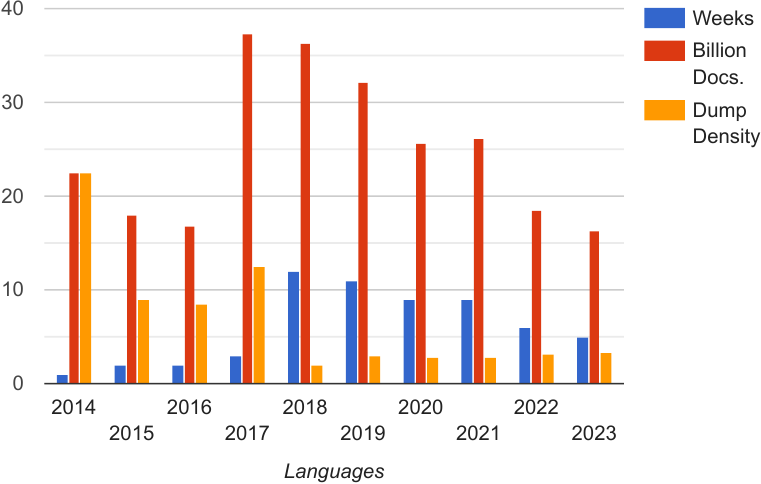}
\caption{Analysis of CommonCrawl dumps used in the dataset, showing the distribution of dump weeks (blue), billion documents (red) and average dump density (yellow) per year.}
\label{fig:dumps_distribution}
\end{figure}

As can be seen from the list, the distribution of dumps across weeks is uneven. Earlier years, particularly 2014 through 2016, have fewer weeks represented compared to more recent years. However, the dumps from these earlier years typically contain a higher dump density, i.e. average document size (see Figure \ref{fig:dumps_distribution}). After 2017, CommonCrawl transitioned to a strategy of producing more frequent dumps, each with relatively smaller average document sizes.

This shift in strategy has important implications for data processing. Special attention must be paid to the earlier years (2014–2016), where deduplication can be more computationally expensive due to the larger document sizes. In these cases, employing stricter filtering mechanisms may be necessary to ensure efficient processing and to avoid excessive resource consumption during deduplication.


\paragraph{Compute power}

Tracking the compute power required during data processing is essential for planning and securing the necessary resources. A well-structured plan not only ensures that adequate resources are available but also provides an opportunity to analyze computational efficiency and sustainability.

Below, we estimated total CPU hours consumed at each stage of the pipeline:


\begin{itemize}
    \item Conversion: On average, converting a dump takes 115,2 CPU hours. For all the data combined this stage took 6,912 CPU hours.
    \item Filtering: On average, filtering one dump takes 763 CPU hours. For all dumps combined, this stage required 45,810 CPU hours. 
    \item Deduplication: Deduplication is notably resource-intensive, taking an average of 3,680 CPU hours per dump. For the entire dataset, this stage consumed 221,230 CPU hours. 
\end{itemize}

Deduplication is by far the most time-consuming step, consuming 80.8\% of the total compute power, which is why it is kept as the final stage in our pipeline (see Section \ref{sec:pipelines.web.dedup}). This is followed by filtering, which takes 16.7\%, and conversion, accounting for 2.5\%.

\subsection{Effects of Filtering and Deduplication}

\begin{figure}
  \centering
  \includegraphics[width=0.75\linewidth]{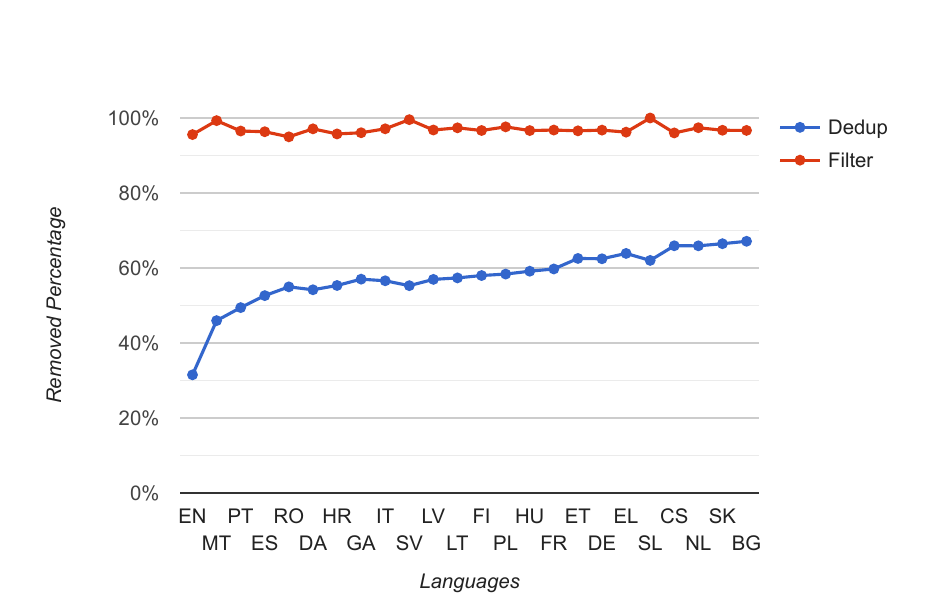}
\caption{Percentage of documents removed after filtering (red) and deduplication (blue) for each language.}
  \label{fig:removed_perc_lang}
\end{figure}

In this section, we examine the effects of filtering and deduplication on the dataset at the document level, focusing on their impact across different languages. These analyses provide insights into how preprocessing steps alter the distribution of available data and highlight potential disparities between languages.

\paragraph{Filtering}

Filtering involves multiple stages, including prefiltering, quality checks, and harmful perplexity filtering, as illustrated in Figure \ref{fig:web_data_pipeline}. By comparing the document sizes before and after filtering, we observe that this step significantly reduces the data for all languages, as shown in Figure \ref{fig:removed_perc_lang}. On average, 96.94\% of the documents are removed during filtering.

Notably, filtering exhibits a moderate positive correlation with the size of the original dataset (r=0.530, p=0.0071). This suggests that larger datasets tend to lose a slightly smaller relative percentage of their content during filtering, but the effect is not uniform across languages. Some languages are disproportionately impacted, potentially due to stricter quality control measures or greater prevalence of low-quality or harmful content in smaller datasets.

\paragraph{Deduplication}


Our deduplication methodology employs a MinHash + Locality Sensitive Hashing (LSH) approach, as detailed in \cite{leveling_helmer_etal2024}, which optimizes precision and recall in identifying and removing duplicate documents. The deduplication step impacts languages differently, with the percentage of removed documents ranging from 31.50\% for English to 67.11\% for Bulgarian (Figure \ref{fig:removed_perc_lang}).

Contrary to intuitive expectations, larger datasets tend to experience a smaller proportion of deduplication. This observation is supported by a strong negative Pearson correlation between the size of the filtered documents and the percentage of data removed during deduplication (r=-0.710, p=0.0001). This suggests that smaller datasets contain a higher proportion of duplicate content relative to their size, which is systematically removed during this stage.

\subsubsection{Disparity Index}

\begin{figure}
\centering
\begin{subfigure}{.5\textwidth}
  \centering
  \includegraphics[width=0.95\linewidth]{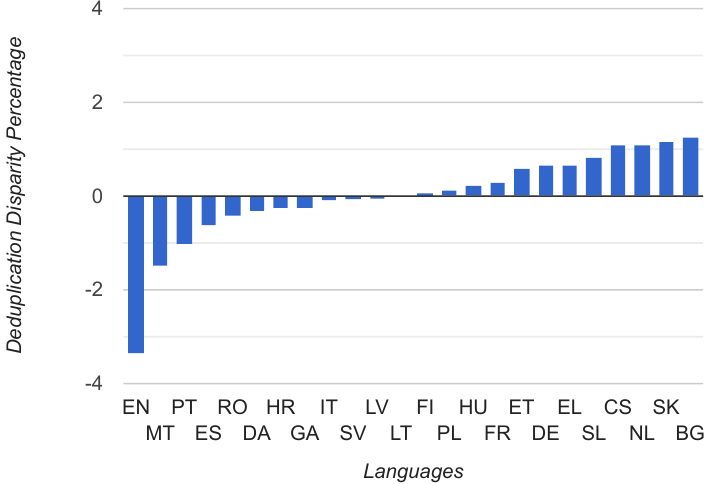}
\caption{Deduplication Disparity Index (DDI).}
  \label{fig:ddi}
\end{subfigure}%
\begin{subfigure}{.5\textwidth}
  \centering
  \includegraphics[width=0.95\linewidth]{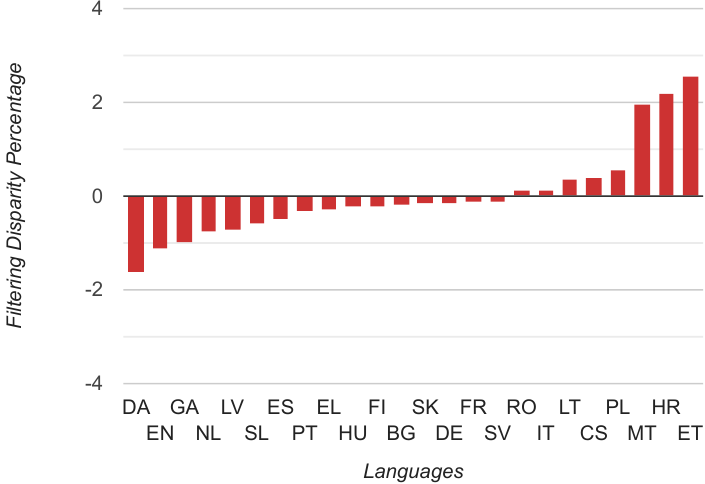}
 \caption{Filtering Disparity Index (FDI).}
\label{fig:fdi}
\end{subfigure}
 \caption{Disparity Indexes for deduplication and filtering, highlighting languages disproportionately affected by each step.} 
\label{fig:disparity_index}
\end{figure}

To quantify the disproportionality in data removal, we introduce the \textit{Disparity Index} (DI). For each language \( l \), the DI is calculated as the Z-score of the removed-to-data ratio \( R_l \) for each language:

\[
\operatorname{DI}_l = \frac{R_l - \mu_R}{\sigma_R}, \quad \text{where} \quad R_l = \frac{r_l}{D_l}
\]

Here, \( r_l \) represents the percentage of documents removed for language \( l \), \( D_l \) is the total number of documents for that language, \( \mu_R \) is the mean of the removed-to-data ratios across all languages, , and \( \sigma_R \) is the standard deviation. This normalization allows for direct comparisons, with positive and negative Z-scores indicating disproportionate removal or retention, respectively.

\paragraph{Filtering Disparity Index (FDI)}

The Filtering Disparity Index (\( \operatorname{FDI} \)) reveals significant variability in how filtering impacts different languages (Figure \ref{fig:fdi}). Languages with a high positive Z-score, such as Estonian (\( \operatorname{FDI}_{et} = 2.554 \)) and Croatian (\( \operatorname{FDI}_{hr} = 2.186 \)), lose a disproportionately large portion of their original data during filtering. Conversely, languages with a high negative Z-score, such as Danish (\( \operatorname{FDI}_{da} = -1.637 \)) and English (\( \operatorname{FDI}_{en} = -1.141 \)), are relatively less affected, indicating a smaller proportion of their content is filtered out.

\paragraph{Deduplication Disparity Index (DDI)}

Similarly, the Deduplication Disparity Index (\( \operatorname{DDI} \)) highlights disparities in deduplication across languages (Figure \ref{fig:ddi}). Languages such as Bulgarian (\( \operatorname{DDI}_{bg} = 1.257 \)) and Slovak (\( \operatorname{DDI}_{sk} = 1.174 \)) are disproportionately impacted, suggesting higher redundancy in their datasets. In contrast, languages such as English (\( \operatorname{DDI}_{en} = -3.373 \)) experience significantly less deduplication, reflecting relatively unique content in their data.


\subsubsection{Summary} 
The web data for this project was sourced from 60 distinct CommonCrawl dumps spanning 2014 2023, resulting in a total of 173 billion documents (703 terabytes of raw data). This broad timeframe captures both historical and contemporary web content.

Deduplication proved to be the most resource-intensive preprocessing step, consuming 80.8\% of the total CPU hours allocated. The introduction of the Deduplication Disparity Index (DDI) and Filtering Disparity Index (FDI) provides a systematic framework for evaluating fairness in the data pipeline. These metrics identify languages disproportionately affected by filtering and deduplication, underscoring the need for tailored processing strategies to ensure equitable data quality and representation across languages.

\section{Insights and Observations}\label{sec:insights}

During the process of data preprocessing,
several challenges emerged that shaped the development of the data pipeline. 
These challenges are important for understanding both the complexity of 
large-scale multilingual dataset processing and the technical, 
organizational, and legal obstacles one can encounter. By presenting these 
observations, we provide key findings and recommendations to guide future 
projects and to navigate similar challenges.

\subsection{Data Quality, Availability and Management}
\label{sec:insights.data}
Data quality and availability are key criteria in the 
construction of datasets for language models. Ensuring high-quality 
data is essential for model performance, while the availability of 
data across different languages, genres, and domains presents unique challenges.

\paragraph{Quality Assessment}
The definition of \textsl{high-quality} data remains a complex and evolving 
topic in the field of LLMs, which has gained more traction as these models 
scale and diversify across languages and domains. Recent trends in dataset 
curation reflect an increasing emphasis on quality over quantity, with 
techniques such as advanced filtering and deduplication becoming more 
prevalent \cite{penedo_kydlicek_etal2024,wang_mrini_etal2024}.

Traditionally, data quality has been assessed through intrinsic signals 
derived from the data itself, such as average word length, sentence complexity, 
or overall structure. However, as the field has advanced, the focus has 
shifted towards leveraging pretrained LLMs to judge high-quality 
content or even generate synthetic data, a strategy that has proven 
effective in improving model performance on specific benchmarks \cite{gunasekar_zhang_etal2023,kumar_choudhary_etal2020,yoo_park_etal2021}. 
One reason for the success of these methods is the alignment between human 
judgments of quality and the assessments made by 
LLMs \cite{wang_liang_etal2023,kocmi_federmann2023,fu_ng_etal2023}.

However, this approach comes with several drawbacks. First, despite the 
promising results, a significant limitation is the lack of transparency in the LLM's mapping between defined criteria and system output in defining \textsl{high-quality} content. There is often 
no clear link between the model’s concept of quality and concrete linguistic 
characteristics, such as sentence structure, readability, or word length. 
Additionally, running LLM inference on large datasets substantially increases 
computational demands, making this approach accessible only to those with 
significant resources. Finally, using LLMs to filter or label data may 
introduce licensing challenges, particularly for commercial applications. We therefore recommend prioritizing more interpretable quality signals 
in data filtering. 

These signals are directly mappable to tangible document 
characteristics, making them transparent and easy to justify. While LLM-based 
filtering techniques show promise, we believe there is a strong need for 
further research into how these models define \textit{high-quality} data. At the 
very least, a hybrid approach combining both traditional and AI-based 
filtering could balance between efficiency, interpretability, 
and scalability.

\paragraph{Availability of Multilingual Data}

As seen in Section \ref{sec:analysis}, one of the primary 
challenges is the significant imbalance in data availability across 
languages. This disparity makes it difficult to create training data 
distributions that ensure downstream models perform well in a wide variety 
of languages. High-resource languages like English dominate the dataset, 
while low-resource languages, such as Irish or Maltese, have far fewer 
available documents, thus requiring special consideration such as avoiding 
filtering and/or deduplicating
to not further reduce the number of tokens available for training in these 
languages.

\paragraph{Versioning and Standardization}

Data versioning plays a crucial role in maintaining a clear record 
of changes and ensuring reproducibility throughout the pipeline. To 
facilitate reproducibility, we opted for a unified format (JSONL) across all 
datasets which makes it easier to track changes and compare versions, 
as any modifications can be identified and versioning information is represented consistently across changes. Furthermore, each step 
of our pipeline saves intermediate results, allowing for a granular 
approach to version control.

In addition to this, the intermediate results were moved 
to remote servers, separate from the training environment, to safeguard 
against data corruption or loss. While this precaution provides an extra 
layer of security, it does come at the cost of increased storage usage, 
which needs to be carefully managed. Nonetheless, data versioning supports tracking changes to the pipeline by storing the corresponding pipeline version in the metadata.

\subsection{Licensing and Compliance}
\label{sec:insights.license}
A significant issue is the lack of clear licensing documentation. 
Researchers sometimes neglect to accurately represent the licenses under which their data falls, leaving legal ambiguities. While license identification remains a complex challenge \cite{SeneviratneKB09,HabernalZG16}, we strongly advocate for more precise legal reporting in dataset publications. Initiatives like the Dataset Cards by Hugging Face\footnote{\url{https://huggingface.co/docs/hub/en/datasets-cards}} are a positive step toward improving licensing transparency. They allow dataset creators to include clear license information by default, promoting better legal clarity. However, it is important to note that adding a license in these dataset cards is not mandatory, which means some datasets may still be published without explicit legal documentation. 

Even when data is available, special care must be taken to comply with the 
licenses under which it is provided. License compliance is crucial from the 
outset, as the licensing terms of any single dataset can impose restrictions 
on the downstream use of merged datasets or models trained on that data. 
Selecting even one dataset with a more restrictive license can constrain 
the licensing flexibility of all derived artifacts, including processed 
datasets and trained models. Ensuring that data sources have compatible and 
permissive licenses is therefore a critical early step in the project.


\subsection{Technical and Organizational Challenges}
\label{sec:insights.organization}

The successful creation of large-scale multilingual datasets involves 
navigating a range of technical and organizational challenges. Addressing 
these issues is essential for maintaining efficiency, ensuring data integrity, 
and fostering effective collaboration among diverse teams. The following 
sections explore the technical challenges of processing data at scale and 
the organizational aspects necessary for project success.

\paragraph{Processing at Scale}

Processing large-scale datasets presents significant technical challenges, 
both in terms of the associated costs and the complexity of managing software, 
hardware, and distributed systems. Large-scale computations often require access 
to multiple computing resources, which introduces additional complexities, 
such as varying system requirements, the need to transfer large volumes of 
data, and the maintenance of software across different platforms. Despite 
the availability of resources, managing distributed processing at scale 
remains a significant barrier, and exploring distributed training approaches 
may offer potential solutions for mitigating some of these difficulties in 
the future.

At the start of any project of this magnitude, it is crucial to consider the 
size of the datasets, the availability of storage and compute resources in 
the clusters, and the allocation of both CPU and GPU resources. One key 
aspect that is often overlooked is ensuring that the processed data is in 
the same location as the final model training environment, as transferring 
large datasets between different clusters can significantly impact efficiency 
and cost. However, budget or resources constraints may limit the ability to 
carry out both data processing and model training on the same cluster. In 
such cases, it becomes essential to carefully plan and optimize across these 
dimensions, considering factors such as storage, computational availability, 
and network transfer costs.

\paragraph{Deduplication Strategy}

In our project, a global deduplication approach (where deduplication is 
applied across multiple CommonCrawl dumps) was found to be impractical 
for two primary reasons. First, for high-resource languages like English, 
the combined dataset before deduplication is too large to be processed 
within our computational environments. Second, as noted in 
\cite{penedo_kydlicek_etal2024}, global deduplication does not provide 
measurable benefits for downstream model performance.

Instead, a local deduplication strategy, applied per-dump and per-language, 
was adopted. This approach reduced the dataset size by approximately 30\%, 
significantly improving processing efficiency in subsequent stages and 
enhancing the overall quality of the training data by minimizing redundancy 
while preserving diversity.

\paragraph{Organizational aspects}
For future large-scale data projects, it is essential to recognize the 
importance of managing diverse expertise, which spans scientific, technical, 
legal, and linguistic domains. Coordinating these competencies is crucial 
for the successful identification and evaluation of data sources, as well 
as for handling the vast volume of data. However, simply collecting data 
is not enough to ensure smooth operations or the sustainability of data 
pipelines and platforms.

We strongly recommend establishing robust data governance and lineage 
strategies at the outset of the project. Defining clear roles, responsibilities, 
and processes early on can significantly improve communication and collaboration 
across teams, aligning efforts as the project evolves. Implementing governance 
frameworks, such as those outlined in \cite{jernite_nguyen_etal2022}, can 
provide valuable structure, helping to maintain efficiency and ensure 
long-term success. Early planning in these areas will prevent organizational 
bottlenecks and provide a foundation for sustainable data management.

\subsection{Adapting to Rapid Innovation}
\label{sec:insights.innovation}

The rapid pace of innovation in data processing methodologies for language 
models introduces a continual challenge. New tools and software frequently 
emerge, offering solutions to existing bottlenecks such as enhanced processing 
efficiency, improved storage capabilities, and advanced filtering techniques. 
However, integrating these innovations into established systems often necessitates 
workflow adaptations, retraining personnel, and potentially reconfiguring entire 
pipelines or rerunning pipelines on all datasets. This process can disrupt 
system stability while requiring considerable investment of time and resources.

Finding a balance between adopting innovations and maintaining operational 
consistency is essential. While the implementation of new software can result 
in performance improvements, it is critical to evaluate which components of 
the system should remain stable to ensure uninterrupted functionality. 
Furthermore, frequent updates (such as newly released curated datasets or 
CommonCrawl dumps) add to the complexity, as these updates necessitate 
continuous modifications to the data pipeline without undermining the 
overall coherence of the system.

A systematic approach to addressing this challenge involves organizing the 
pipeline into modular components from the outset. By clearly separating and 
labeling each segment of the pipeline, it becomes possible to identify which 
elements are adaptable to future innovations and which should remain static due 
to their critical nature. For instance, components such as deduplication processes
may benefit from frequent optimization, whereas core elements like metadata 
normalization or key filtering mechanisms might need to remain constant to 
maintain reliability. Furthermore, depending on the stage of the project, it 
may be prudent to commit to a particular solution, even if suboptimal, rather 
than continuously reconfiguring the system.

\section{Conclusion and Outlook}\label{sec:conclusion}

In this paper, we presented the entire process of data preparation for the OpenGPT-X project. We outlined the requirements that guided our 
data selection, which conceptually split our sources into two categories: curated data and web data. This distinction shaped our approach to the implementation of our data pipelines: one for curated data, which required minimal filtering, and another 
for web data, which focused heavily on filtering and deduplication. In addition to providing 
a thorough description of our pipeline and data preparation methods, we included an 
in-depth analysis of the resulting datasets. 
This analysis ensures transparency, aligning with European data recommendations and 
best practices in scientific research.

Finally, to contribute to the broader research field and promote openness, we 
highlighted the challenges faced during this project and the lessons learned. By sharing these insights, we aim to provide valuable guidance for future projects that undertake large-scale data preparation for multilingual  language models.


As part of our future research, we plan to investigate the use of 
synthetic data generation for the creation of LLM pretraining data 
and for advanced data quality filtering, e.g. based on LLMs as a judge.
Furthermore, we are looking into generating training data that is compliant
with current or upcoming laws and regulations (EU AI Act, GDPR), e.g.
by removing or masking personally identifiable information.
We intend to extend our collection of datasets, also targeting
both datasets for underrepresented domains as well as languages.
With additional processing requirements due to 
advanced processing techniques,
and a growing number of available datasets, 
the demand for compute resources to generate high-quality training data
will also increase.



\section*{Acknowledgements}
This work was funded by the German Federal Ministry for Economic Affairs 
and Climate Action (BMWK) through the project OpenGPT-X (project
no. 68GX21007D).
The authors gratefully acknowledge the Gauss Centre 
for Supercomputing e.V. (\url{http://www.gauss-centre.eu})
for providing compute resources
on the GCS Supercomputer JUWELS at Jülich
Supercomputing Centre (JSC).

The authors gratefully acknowledge the compute resources made available to them on the high-performance computer at the NHR Center of TU Dresden. This center is jointly supported by the Federal Ministry of Education and Research and the state governments participating in the NHR (\url{http://www.nhr-verein.de/unsere-partner}).

Many thanks to the OpenGPT-X project partner IONOS for providing a NVIDIA DGX-2 machine 
to facilitate development of our data pipelines as well as processing
many curated datasets.


We acknowledge the use of AI tools (i.e. ChatGPT-4/Claude) in the early stages  of this paper
for improving readability and brainstorming useful insights for the analysis section. 

\bibliographystyle{plainurl}
\bibliography{main}
\pagebreak
\addtocounter{section}{1}

\section*{Appendix}\label{sec:appendix}

\subsection{Dataset}


\begin{table}[htb]
\centering
\resizebox{0.99\textwidth}{!}{%
\begin{tabular}{lllll}
\hline
\multicolumn{1}{l}{\textbf{Monolingual Dataset}}                                 
& \multicolumn{1}{l}{\textbf{Text Extraction}} 
& \multicolumn{1}{l}{\textbf{Language Identification}} 
& \multicolumn{1}{l}{\textbf{Filtering}}                               
& \textbf{Deduplication}                        
\\ \midrule

\multicolumn{1}{l}{C4 \cite{raffel_shazeer_etal2020}}                    
& \multicolumn{1}{l}{CC (WARC)}                
& \multicolumn{1}{l}{langdetect  ($\geq 0.99$)}          
& \multicolumn{1}{l}{Non-language, outliers, LID}                      
& 3-sentence spans                              \\
\multicolumn{1}{l}{The Pile \cite{gao_biderman_etal2020}}                      
& \multicolumn{1}{l}{jusText}                  
& \multicolumn{1}{l}{pycld2, fastText}                 
& \multicolumn{1}{l}{LID}                                              
& MinHashLSH                                    \\
\multicolumn{1}{l}{RefinedWeb \cite{penedo_malartic_etal2023}}           
& \multicolumn{1}{l}{trafilatura}              
& \multicolumn{1}{l}{fastText}                         
& \multicolumn{1}{l}{LID, DW/ LW heuristics}                           
& MinHashLSH, token matching, URL deduplication \\
\multicolumn{1}{l}{Dolma \cite{soldaini_kinney_etal2024}}                   
& \multicolumn{1}{l}{CC (WARC)}                
& \multicolumn{1}{l}{fastText  ($\geq 0.5$)}             
& \multicolumn{1}{l}{LID , DW/ L, Toxic content}                       
& URL matching, raw doc match, Bloom filter     \\
\multicolumn{1}{l}{FineWeb \cite{penedo_kydlicek_etal2024}}                 
& \multicolumn{1}{l}{trafilatura}              
& \multicolumn{1}{l}{fastText}                         
& \multicolumn{1}{l}{DW/ LW heuristics}                                
& MinHash                                       
\\ 
\midrule
\multicolumn{1}{l}{\textbf{Multilingual Dataset}}                                 
& \multicolumn{1}{l}{\textbf{Text Extraction}} 
& \multicolumn{1}{l}{\textbf{Language Identification}} 
& \multicolumn{1}{l}{\textbf{Filtering}}                               
& \textbf{Deduplication}       \\ 
\midrule
\multicolumn{1}{l}{CCNet \cite{wenzek_lachaux_etal2019}}                     
& \multicolumn{1}{l}{WET}                      
& \multicolumn{1}{l}{fastText ($\geq 0.5$)}              
& \multicolumn{1}{l}{Perplexity filtering, LID}                        
& PW SHA-1                                      \\
\multicolumn{1}{l}{mC4 \cite{xue_constant_etal2021}}                                
& \multicolumn{1}{l}{CC (WARC)}                
& \multicolumn{1}{l}{cld3}                             
& \multicolumn{1}{l}{C4, 3 lines of 200+ characters}                   
& 3-sentence spans                                           \\
\multicolumn{1}{l}{OSCAR 22.01 \cite{abadji_suarez_etal2022}}             
& \multicolumn{1}{l}{WET}                      
& \multicolumn{1}{l}{fastText ($\geq 0.8$)}              
& \multicolumn{1}{l}{LW LID,  WD Unicode rules, UT1 blocklist}         
& Line-wise deduplication (English only)        \\
\multicolumn{1}{l}{BigScience ROOTS \cite{laurenccon_saulnier_etal2022}} 
& \multicolumn{1}{l}{Custom extractor}         
& \multicolumn{1}{l}{Manual (by data source)}      
& \multicolumn{1}{l}{Word frequency heuristics}                        
& SimHash, substring deduplication              \\
\multicolumn{1}{l}{Glot500 \cite{imanigooghari_lin_etal2023}}          
& \multicolumn{1}{l}{Custom crawling}          
& \multicolumn{1}{l}{Manual (by data source)}                               
& \multicolumn{1}{l}{Language-script matching, BigScience ROOTS rules} & Sentences                                           \\
\multicolumn{1}{l}{RedPajama-v2 \cite{redpajama2023}}        
& \multicolumn{1}{l}{WET}           
& \multicolumn{1}{l}{fastText}  
& \multicolumn{1}{l}{Classifier, heuristic filtering}                  
& DW Bloom filter                               \\
\multicolumn{1}{l}{MADLAD-400 \cite{kudugunta_caswell_etal2024}}            
& \multicolumn{1}{l}{Unknown}                      
& \multicolumn{1}{l}{Semi-supervised LID}              
& \multicolumn{1}{l}{Prefiltering similar to C4}                       
& LW deduplication                              \\
\multicolumn{1}{l}{HPLT \cite{degilbert_nail_etal2024}}                            
& \multicolumn{1}{l}{warc2text}                
& \multicolumn{1}{l}{CLD2, fastText}                   
& \multicolumn{1}{l}{Two-stage LID, dictionary spell check}            
& DW MinHash                                    \\ \hline
\end{tabular}%
}
\caption{\label{tab:datasets}%
This table summarizes the text extraction, language identification, filtering, and deduplication methods used across various large-scale datasets discussed in the related work section (Section \ref{sec:rel_work}). It provides a comparison of the approaches taken for both monolingual and multilingual datasets, illustrating the diversity of tools and techniques employed in the preprocessing of data for language model training. Abbreviations: DW = Document-wise, LW = Line-wise, PW = Paragraph-wise, LID = Language Identification.}

\end{table}

\subsection{Metadata Content Example}
\label{ssec:appendix.metadataexample}

\begin{verbatim}
    {
    "meta": {
        "url": "https://en.wikipedia.org/wiki/Organic%20Chemistry/Cover", 
        "title": "Organic Chemistry/Cover", 
        "docid": "wikimedia/wikibooks/enwikibooks-20240401", 
        "language": "en", 
        "language_score": 1.0, 
        "download_data": "2024-04-01", 
        }
    "text": "Welcome to the world's foremost open content
            Organic Chemistry Textbook on the web!\n\n
            The Study of Organic Chemistry[...]", 
    }
\end{verbatim}

\subsection{Curated Data}
\label{sec:appendix.curated}

\subsubsection{Filtering}
\label{sec:appendix.curated.filtering}
The correlation between the percentage filtered and the total number of words in a dataset was relatively low at \( r = 0.26 \) (p = 0.022), suggesting only a weak relationship. Similarly, a slightly stronger correlation was observed between the percentage filtered and the number of documents, at \( r = 0.33 \) (p = 0.003), indicating that datasets with more documents tend to require somewhat more filtering, but the relationship is still far from definitive.

We also examined the impact of format on filtering, but found no significant correlation (\( r = 0.03 \), p = 0.79). However, when considering the language of the dataset, we found a more notable correlation (\( r = 0.37 \), p = 0.001), hinting that certain languages may be more prone to needing extensive filtering, possibly due to differences in data quality or availability across languages.

Finally, domain showed no significant correlation with the percentage of data filtered (\( r = -0.004 \), p = 0.97).

These results suggest that filtering operates somewhat independently from these broader dataset characteristics. Language differences, in particular, seem to warrant more focused attention in future analyses, as they exhibit a stronger connection to filtering needs compared to other factors. 

\subsubsection*{List of Curated Datasets}
The complete list of curated datasets is presented in Table \ref{tab:curated_data_list} (next page).

\newgeometry{top=1cm,bottom=1.5cm,left=0.5cm,right=0.5cm}

\small
\renewcommand{\arraystretch}{1.5} 
\begin{longtable}{p{4cm}p{3cm}p{2cm}p{2.7cm}p{2.5cm}>{\raggedleft\arraybackslash}p{1.7cm}>{\raggedleft\arraybackslash}p{1.5cm}>{\centering\arraybackslash}p{1.3cm}}
\caption{\label{tab:curated_data_list}%
This table provides an overview of various multilingual datasets utilized in OpenGPT-X project. Each entry begins with the \textbf{Corpus ID} of the dataset and a link to its project page. The \textbf{Language(s)} column specifies the languages included within each dataset (see Section \ref{sec:analysis.curated}). The \textbf{Format} of the datasets is also noted, indicating the file type, such as TXT, JSON, or XML. The \textbf{License} column outlines the legal terms governing the use of each dataset, for uncommon licenses a link is provided. The \textbf{Domain} column reflects the specific field or subject area that the dataset pertains to, such as Law, Math, or Medical. The  \textbf{\#Docs} column presents the total number of documents contained in each dataset, while the \textbf{\#Words} column conveys the total word count in thousand. }\\%
\toprule
\textbf{Corpus ID} &
  \textbf{Language(s)} &
  \textbf{Format} &
  \textbf{License} &
  \textbf{Domain} &
  \textbf{\#Docs} &
  \textbf{\#Words} \\  \midrule 
\endfirsthead
\multicolumn{8}{c}%
{{\bfseries Table \thetable\ continued from previous page}} \\ \hline
\textbf{Corpus ID} &
  \textbf{Language(s)} &
  \textbf{Format} &
  \textbf{License} &
  \textbf{Domain} &
  \textbf{\#Docs} &
  \textbf{\#Words}  \\  \midrule 
\endhead%
\href{https://dcl.bas.bg/BulNC-registration/?lang=EN}{eu\_admin} & EU24 & TXT & CC BY-NC 3.0 & Law \& Admin. & 435.318 & 1.079  
\\
\href{https://github.com/hendrycks/math}{ampsmath\_khan} \cite{hendrycks_burns_etal2021} & EN & JSON / TXT & MIT & Math & 102.985 & 12.081  
\\
\href{https://github.com/hendrycks/math}{ampsmath\_mathematica} \cite{hendrycks_burns_etal2021} & EN & JSON / TXT & MIT & Math & 4.824.189 & 232.502  
\\
\href{https://zenodo.org/records/4643066}{cdrs\_bt} \cite{fobbe_2021_4643066} & DE & TXT & CC0 1.0 & Law \& Admin. & 124.487 & 515.053  
\\
\href{https://zenodo.org/records/4006645}{ce\_bag} \cite{fobbe_2020_4006645} & DE & TXT & CC0 1.0 & Law \& Admin. & 5.290 & 18.487  
\\
\href{https://zenodo.org/records/7691841}{ce\_bfh} \cite{fobbe_2023_7691841} & DE & TXT & CC0 1.0 & Law \& Admin. & 10.310 & 22.635  
\\
\href{https://zenodo.org/records/7699032}{ce\_bgh} \cite{fobbe_2023_7699032} & DE & TXT & CC0 1.0 & Law \& Admin. & 71.976 & 104.002  
\\
\href{https://zenodo.org/records/7767295}{ce\_bpatg} \cite{fobbe_2023_7767295}& DE & TXT & CC0 1.0 & Law \& Admin. & 30.772 & 66.029  
\\
\href{https://zenodo.org/records/5910152}{ce\_bverfg} \cite{fobbe_2022_5910152} & DE & TXT & CC0 1.0 & Law \& Admin. & 8.449 & 20.679  
\\
\href{https://zenodo.org/records/7749683}{ce\_bverwg} \cite{fobbe_2023_7749683} & DE & TXT & CC0 1.0 & Law \& Admin. & 27.099 & 49.664  
\\
\href{https://zenodo.org/records/4542662}{cpp\_bt}  \cite{fobbe_2021_4542662}& DE & TXT & CC0 1.0 & Law \& Admin. & 4.087 & 235.502  
\\
\href{https://www.gaois.ie/en/corpora/monolingual}{gaois\_corpus} & GA & XML / TMX & CC BY 4.0 & Law \& Admin. & 2 & 2.168  
\\
\href{https://zenodo.org/records/10030647}{cd\_icj} \cite{fobbe_2023_10030647} & EN, FR & TXT & CC0 1.0 & Law \& Admin. & 4.565 & 25.979  
\\
\href{https://zenodo.org/records/7051934}{cd\_pcij} \cite{fobbe_2022_7051934}  & EN & TXT & CC0 1.0 & Law \& Admin. & 518 & 2.117  
\\
\href{https://gigaword.dk/}{dagw} \cite{dercynski_closici_etal2021}  & DA & TXT & CC BY 4.0 & Web & 285.634 & 922.700  
\\
\href{https://www.deutschestextarchiv.de/}{dta} & DE & TXT & CC BY-SA 4.0 & Books & 5.305 & 197.339  
\\
\href{https://joint-research-centre.ec.europa.eu/language-technology-resources/dcep-digital-corpus-european-parliament_en}{dcep} \cite{hajlaoui_kolovratnik_etal2014} & EU24 & TXT & \href{https://commission.europa.eu/content/european-union-public-licence_en}{EUPL}  & Law \& Admin. & 1.227.996 & 1.117  
\\
\href{https://pub.cl.uzh.ch/wiki/public/pacoco/medi-notice}{medi\_notice} \cite{graen_kew_etal2019} & DE, FR, IT & TSV & CC BY SA 
& Medical & 22.219 & 52.068  
\\
\href{https://www.sketchengine.eu/estonian-national-corpus/#:~:text=Estonian\%20National\%20Corpus\%202021\%20(Estonian\%20NC\%202021)\%20\%E2\%80\%93\%202.4\%20billion,academic\%20writing\%20(2020\%E2\%80\%932021)}{enc2021} \cite{koppel_kallas_2022} & EU24 & VERT & CC BY-NC 4.0 & Web & 4.361.539 & 1.004  
\\
\href{https://www.cl.ut.ee/korpused/segakorpus/}{estonian\_reference\_corpus} & ET & XML & 
CC BY-NC\textsuperscript{\textdagger} 
& Web & 17.892 & 227.978  
\\
\href{https://opus.nlpl.eu/EMEA/corpus/version/EMEA}{emea} \cite{tiedemann2012parallel} & EU24 & TXT & CC BY 4.0 & Medical & 17.960 & 152.252  \\
\href{https://www.statmt.org/europarl/}{europarl} \cite{koehn2005} & EU24 & XML & CC0 1.0\textsuperscript{\textdagger} & Law \& Admin. & 63.937 & 660.716  
\\
\href{https://eur-lex.europa.eu/homepage.html?locale=en}{eurlex} \cite{laurer_borrett2020} & EU24 & JSONL & CC BY 4.0 & Law \& Admin. & 4.463.480 & 13.235  
\\
\href{https://opus.nlpl.eu/ECB/corpus/version/ECB}{ecb\_corpus} \cite{tiedemann2012parallel} & EU24 & XML & CC0 & Finance & 19 & 53.894   
\\
\href{https://github.com/nkrusch/fi-news-corpus}{fi\_news} & FI & CSV & MIT & News & 69.413 & 3.650 
\\
\href{https://openlegaldata.io/research/2019/02/19/court-decision-dataset.html}{german\_legal\_cases} \cite{ostendorff_2020_towards} & DE & JSONL & CC0 1.0\textsuperscript{\textdagger} & Law \& Admin. & 249.240 & 749.060  
\\
\href{https://doi.org/10.5281/zenodo.3611246}{german\_political\_speeches} \cite{adrien_barbaresi_2020_3611246} & DE & XML & CC BY-SA 4.0 & Law \& Admin. & 6.659 & 11.366  
\\
\href{https://huggingface.co/datasets/greek_legal_code}{greek\_legal\_code} \cite{papaloukas_etal_2021_glc} & EL & JSONL & CC BY 4.0 & Law \& Admin. & 40.929 & 28.791  
\\
\href{https://www.gaois.ie/en/corpora/parallel/data/}{irish\_legislation} & GA & XML / TMX & CC BY 4.0 & Law \& Admin. & 12 & 29.136  
\\
\href{https://kleineanfragen.de/info/daten}{kleine\_anfragen} & DE & SQL & \href{https://opendatacommons.org/licenses/odbl/1-0/}{ODbL 1.0} & Law \& Admin. & 3.401.475 & 6.151 
\\
\href{https://huggingface.co/datasets/MLRS/korpus_malti}{korpus\_malti} \cite{BERTu} & MA & JSONL & CC BY-NC-SA & Web & 103.874 & 276.204 
\\
\href{https://huggingface.co/datasets/joelniklaus/legal-mc4}{legal\_mc4} & EU24 & JSONL & CC BY 4.0 & Law \& Admin. & 4.853.498 & 13.892  
\\
\href{https://marcell-project.eu/}{marcell} \cite{varadi_koeva_etal2020} & BG, HR, HU, RO, SL, SK, PL & CONLLUP / TXT & CC0 1.0 & Law \& Admin. & 280.629 & 373.283  
\\
\href{https://macocu.eu/}{macocu} \cite{banon_esplagomis_etal2022} & EU24 & XML & CC0 1.0 & Web & 79.164.225 & 27.930  \\
\href{https://www.euromatrixplus.net/multi-un/}{multi\_un} \cite{eisele_chen2010} & EU24 & XML & CC BY-NC\textsuperscript{\textdagger} & Law \& Admin. & 262.391 & 1.096  
\\
\href{https://huggingface.co/datasets/coastalcph/multi_eurlex#:~:text=MultiEURLEX\%20comprises\%2065k\%20EU\%20laws,\%2C\%20\%5B1115\%2C\%20fruit\%5D.}{multi\_eurlex} \cite{aueb2021} & EU24 & JSONL & CC BY-NC-SA 4.0 & Law \& Admin. & 1.038.305 & 1.184  
\\
\href{https://huggingface.co/datasets/HiTZ/Multilingual-Medical-Corpus}{medical\_t5} \cite{garcíaferrero2024medical} & DE, EN, ES, FR & JSON & Apache-2.0 & Medical & 94.269 & 90.741  
\\
\href{https://huggingface.co/datasets/NbAiLab/NCC}{ncc} \cite{kummervold_etal_2021_operationalizing} & NO & JSON & \href{https://huggingface.co/datasets/NbAiLab/NCC#license}{Various}\footnote{The ncc corpus includes the following licenses : NLOD 2.0, CC0 1.0, CC BY-NC 2.0, CC BY-SA 3.0} & Knowledge Base & 1.241.694 & 2.920 
\\
\href{http://openlegaldata.io/research/2019/02/19/court-decision-dataset.html}{openlegaldata} \cite{ostendorff_2020_towards} & DE & JSONL & CC0 1.0 & Law \& Admin. & 103.870 & 363.962  
\\
\href{http://www.opensubtitles.org/}{opensubtitles2018} \cite{lison_tiedemann2016} & EU24 & XML & CC BY-NC\textsuperscript{\textdagger} & Culture & 2.257.513 & 1.398 
\\
\href{https://huggingface.co/datasets/open-web-math/open-web-math}{open\_web\_math} \cite{paster2023openwebmath} & EN & Parquet & ODC-By 1.0 & Math & 5.956.611 & 7.378  
\\
\href{https://www.corpusitaliano.it/en/}{paisa} \cite{lyding_stemle_etal2014} & IT & TXT / XML & CC BY-NC-SA 3.0 & Web & 318.730 & 207.382  
\\
\href{https://www.clarin.eu/parlamint}{parlamint} \cite{erjavic_etal2022} & EU24 & XML / TEI & CC BY & Law \& Admin. & 40.995 & 1.095  
\\
\href{https://huggingface.co/datasets/allenai/peS2o}{pes2o} \cite{soldaini_lo2023} & EN & JSONL & ODC-By 1.0 & Academic & 38.958.175 & 42.172  
\\
\href{https://pile.eleuther.ai/}{pile\_v2\_freelawopinions} \cite{pile} & EN & JSONL & CC BY-ND 4.0 & Law \& Admin. & 4.410.012 & 10.408  
\\
\href{https://github.com/EleutherAI/hn-scraper}{pile\_hackernews} \cite{pile} & EN & JSONL & CC BY-NC & Web & 331.205 & 246.149  
\\
\href{https://pile.eleuther.ai/}{pile\_nih\_exporter} \cite{pile} & EN & JSONL & CC0 1.0 & Medical & 930.048 & 307.055  
\\
\href{https://pile.eleuther.ai/}{pile\_openwebtext2} \cite{pile} & EN & JSONL & MIT\textsuperscript{\textdagger} & Books & 15.752.217 & 10.633  
\\
\href{https://pile.eleuther.ai/}{pile\_v2\_philarchive} \cite{pile} & EN & JSONL & Various\footnote{This Pile part is extracted from \href{https://arxiv.org/}{ArXiv} where the author can decide between a \href{https://info.arxiv.org/help/license/index.html}{variety of licenses}.} & Academic & 40.344 & 487.032  
\\
\href{https://pile.eleuther.ai/}{pile\_pmc\_abstracts} \cite{pile} & EN & JSONL & Various\footnote{This Pile part is extracted from \href{https://pubmed.ncbi.nlm.nih.gov/}{PubMed} where the author can decide between a \href{https://www.ncbi.nlm.nih.gov/pmc/about/copyright/}{variety of licenses}.}& Medical & 15.476.085 & 3.173  
\\
\href{https://pile.eleuther.ai/}{pile\_pmc\_extracts} \cite{pile} & EN & JSONL & Various\footnote{This Pile part is extracted from \href{https://pubmed.ncbi.nlm.nih.gov/}{PubMed} where the author can decide between a \href{https://www.ncbi.nlm.nih.gov/pmc/about/copyright/}{variety of licenses}.} & Medical & 2.808.849 & 12.113  
\\
\href{https://www.sketchengine.eu/polish-parliamentary-corpus/#:~:text=The\%20Polish\%20Parliamentary\%20Corpus\%20(PPC,segments\%20of\%20interpellations\%20and\%20questions}{ppc} \cite{ogrodniczuk2018} & PL & XML & CC BY 4.0 & Law \& Admin. & 38.304 & 472.193  
\\
\href{https://elrc-share.eu/repository/search/?q=PRINCIPLE&selected_facets=languageNameFilter_exact\%3AIrish}{principle\_ga}  & TXT & GA, EN & CC BY 4.0 & Law \& Admin. & 11 & 25.776.098   
\\
\href{https://www.projekt-gutenberg.org/}{projekt\_gutenberg}& EU24 & TXT & \href{https://www.gutenberg.org/policy/license.html}{Custom Licence} & Books & 60.912 & 3.374  
\\
\href{https://huggingface.co/datasets/hoskinson-center/proof-pile}{proof\_pile} & EN & JSONL & Apache-2.0 & Math & 2.128.218 & 4.495  
\\
\href{https://github.com/togethercomputer/RedPajama-Data}{rp\_arxiv} \cite{together2023redpajama} & EN & JSONL & Apache-2.0 & Academic & 1.503.469 & 10.202  
\\
\href{https://opus.nlpl.eu/legacy/SETIMES.php}{setimes\_corpus} \cite{tiedemann2012} & BG, BS, EL, EN, HR, MK, RO, SQ & TXT & CC BY-SA 3.0 & News & 246.206 & 69.279  
\\
\href{https://data.europa.eu/data/datasets/elrc_1188?locale=en}{seimas} \cite{morkevicius_briediene_etal2021} & EN, LT & XML / TMX & CC BY 4.0 & Law \& Admin. & 3540 & 47.319  
\\
\href{https://www.juls.savba.sk/justicecorp.html}{sk\_court\_decisions} & SVK & JSONL & CC0 1.0\textsuperscript{\textdagger} & Law \& Admin. & 1.647.736 & 2.174 
\\
\href{http://nlp.ffzg.hr/resources/corpora/slwac/}{slwac} \cite{ljubesic_erjavec2011} & SL & CONLLU & CC BY-SA 4.0 & Web & 1.484.546 & 953.803 
\\
\href{https://zenodo.org/records/5495529}{spanish\_legal\_corpora} \cite{gutierrezfandino_armengolestape_etal2021} & ES & TXT & CC BY 4.0 & Law \& Admin. & 15 & 1.383.749  
\\
\href{https://archive.org/details/stackexchange}{rp\_stackexchange} & EN & JSONL & CC BY-SA 4.0 & Forum & 23.909.364 & 7.311 
\\
\href{https://huggingface.co/datasets/bigcode/starcoderdata}{starcoder\_data} \cite{li_allal_etal2023} & XX & JSONL & Various\footnote{The Stack is a collection of source code from repositories with various licenses. Any use of all or part of the code gathered in The Stack must abide by the terms of the original licenses, including attribution clauses when relevant.} & Source Code & 206.634.734 & 73.064  
\\
\href{https://huggingface.co/datasets/rcds/swiss_judgment_prediction}{swiss\_judgment\_prediction} \cite{niklaus2022empirical} & DE, FR & JSONL & CC BY-SA 4.0 & Law \& Admin. & 262.789 & 144.599  
\\
\href{https://pub.cl.uzh.ch/wiki/public/pacoco/swiss_legislation_corpus}{swiss\_legislation\_corpus} \cite{hoefler_piotrowski2011} & DE, FR & TXT & CC BY-SA & Law \& Admin. & 2 & 10.407  
\\
\href{https://opendata.swiss/en/dataset?keywords_en=environmental--policy&keywords_en=policy-analysis}{swiss\_policy\_documents} \cite{graen_kew_etal2019} & DE, FR, IT & PARQUET & CC BY 4.0 & Law \& Admin. & 417.923 & 561.574  
\\
\href{https://wacky.sslmit.unibo.it/doku.php?id=start}{wacky} \cite{baroni_bernardini_etal2009} & EN, DE, IT & TXT & CC BY-NC-SA 4.0 & Web & 8.571.072 & 6.440  
\\
\href{https://ucsb.box.com/s/ap23l8gafpezf4tq3wapr6u8241zz358}{wikihow} \cite{koupaee_wang2018} & EN & CSV & CC BY-NC-SA & Knowledge Base & 210.526 & 109.611  
\\
\href{https://en.wikibooks.org/wiki/Main_Page}{wikibooks} \cite{wikimedia} & EU24 & XML & CC BY-SA 4.0 & Books & 190.332 & 168.473  
\\
\href{https://en.wikinews.org/wiki/Main_Page}{wikinews} \cite{wikimedia} & EU24 & XML & CC BY 2.5 & News & 197.265 & 46.936  
\\
\href{https://en.wikipedia.org/wiki/Main_Page}{wikipedia} \cite{wikimedia} & EU24 & XML & CC BY-SA 4.0  & Knowledge Base & 26.175.600 & 11.382  
\\
\href{https://www.wikiquote.org/}{wikiquote} \cite{wikimedia} & EU24 & XML & CC BY-SA 4.0 & Recreation & 208.881 & 146.497  
\\
\href{https://en.wikisource.org/wiki/Main_Page}{wikisource} \cite{wikimedia} & EU24 & XML & CC BY-SA 4.0 & Books & 428.157 & 798.322  
\\
\href{https://en.wikivoyage.org/wiki/Main_Page}{wikivoyage} \cite{wikimedia} & EU24 & XML & CC BY-SA 4.0 & Culture & 71.905 & 66.278  
\\
\href{https://www.wikiversity.org/}{wikiversity} \cite{wikimedia} & EU24 & XML & CC BY-SA 4.0 & Books & 61.397 & 65.482 
\\ \bottomrule 
\end{longtable}



\end{document}